\documentclass{article}

\PassOptionsToPackage{numbers, sort&compress}{natbib}
\usepackage[main, final]{neurips_2026}

\usepackage[utf8]{inputenc}
\usepackage[T1]{fontenc}
\usepackage{hyperref}
\usepackage{url}
\usepackage{booktabs}
\usepackage{amsfonts}
\usepackage{nicefrac}
\usepackage{bm}
\usepackage{microtype}
\usepackage{xcolor}
\usepackage{caption}
\usepackage{graphicx}
\usepackage{amsmath}
\usepackage{multirow}
\usepackage{changepage}
\usepackage{rotating}
\usepackage{pdflscape}
\hypersetup{hidelinks}
\usepackage{caption}

\title{Measuring Robustness and Efficiency in a Connectome-Constrained Fly Visual System Model: Linking an Adversarial and a Perceptual Illusion}

\author{%
  Alexandre Garcia-Duran\textsuperscript{1,2,*} \quad
  Alejandro Rodriguez-Garcia\textsuperscript{3,*} \\
  \textbf{Manuel Molano-Mazón}\textsuperscript{1, 2}
  \quad
  \textbf{Alexandre Hyafil}\textsuperscript{1} \quad
  \textbf{Srikanth Ramaswamy}\textsuperscript{3} \\
  \textbf{Janne K. Lappalainen}\textsuperscript{4} \\[4pt] 
  \textsuperscript{1}Centre de Recerca Matemàtica, Barcelona, Spain \\
  \textsuperscript{2}Universitat Politècnica de Catalunya, Barcelona, Spain \\
  \textsuperscript{3}Newcastle University, Newcastle upon Tyne, United Kingdom \\
  \textsuperscript{4}Allen Institute, Seattle, WA, United States \\
  \texttt{janne.lappalainen@alleninstitute.org} \\
  [2pt]
  \textsuperscript{*}These authors contributed equally. \quad
}

\begin{document}

\maketitle

\begin{abstract}


Biological visual systems compute reliably under tight metabolic and wiring constraints, whereas artificial vision systems consume far more energy and can be sensitive to adversarial perturbations. Whether networks with brain-like wiring are more robust at lower computational cost is unclear. We address this by benchmarking a task-optimized, continuous-time recurrent neural network of the fly visual system constrained by its synaptic connectome (FlyNet) against artificial neural networks. We compare it against CNNs, RAFT, and continuous-time recurrent neural networks without detailed connectome constraints (sCTRNN, CTRNN) on a collision-detection task within a differentiable 3D rendering pipeline. FlyNet is adversarially vulnerable but, relative to its compute, more robust to common video corruptions than RAFT, while using up to five orders of magnitude fewer FLOPs. The CTRNNs remain more robust than FlyNet across almost all tested perturbations, suggesting that connectome constraints benefit efficiency, through their low edge count, over robustness. Recurrence improves robustness to white-box attacks in RAFT and temporal context improves robustness to common corruptions.
FlyNet's position and velocity representations degrade under adversarial textures but recover under counterfactual textures that restore correct classification while staying equally far from training, ruling out unfamiliar textures as the cause of failure. The reverse-phi illusion similarly disrupts these representations and inverts the same low-level motion statistic as the adversarial textures. Optimizing this input statistic alone, without access to the network, suffices to induce classification failure, attributing both perturbations to the same circuit vulnerability. Connectome-constrained models can thus surface adversarial failure mechanisms as testable hypotheses about biological perceptual vulnerabilities.

\end{abstract}

\section{Introduction}

Biological neural systems compute reliably despite stochastic components and variable inputs \citep{Faisal2008-cj}, and do so under tight metabolic and wiring constraints \citep{Attwell2001-ia, Laughlin1998}. But these systems are not perturbation-proof: Classical visual illusions expose the limits of computational mechanisms of biological perception \citep{Eagleman2001-mh}. Artificial neural networks, by contrast, consume vastly more energy \citep{Patterson2021-fe} and can fail under various perturbations \citep{szegedy2014, goodfellow2015, Madry2018, Hendrycks2019-ez}.
%
%
Do artificial and biological neural networks specialized for the same task share the same failure modes? Linking both has been hard because the systems one can attack with gradients are not biologically wired. To our knowledge, no perceptual illusion has yet been tied to an adversarial stimulus in the same system at the level of a computation both exploit.

Robustness --- the ability to maintain reliable computation under perturbations --- of a biological neural network can now directly be evaluated with FlyNet \citep{Lappalainen2024}, a fly visual system model constrained by real synaptic wiring. It was trained on motion detection, an ecologically essential task for the fly, using the Sintel optical flow dataset \citep{Butler_2012} and recapitulates single-neuron response properties across three decades of experimental research, without training on that data. In artificial networks, robustness is linked to specific architectural features --- recurrence \citep{Kar2019-da, Spoerer2020-em, greco2026}, global processing \citep{Geirhos2018-zd, Naseer2021-hd}, and continuous-time dynamics \citep{Yan2019-sf, Dapello2020-hv}. FlyNet has little of these: it uses continuous-time dynamics but has sparse recurrence and primarily local connectivity, reflecting the predominantly feedforward organization of the fly optic lobe. Any robustness it exhibits might therefore arise from properties other than recurrence or global integration, possibly shaped through evolution under tight metabolic constraints \citep{Attwell2001-ia, Laughlin1998}. Does biological wiring contribute to robustness, to efficiency, or to both?

We designed a behaviorally grounded, looming-based collision-detection task to evaluate robustness in both connectome-constrained networks and conventional optic-flow models. We implemented the task in a fully differentiable 3D rendering pipeline, in which textured objects follow parametric trajectories toward the observer. Rather than optimizing perturbations in pixel-space --- where they lose effectiveness under geometric transformations \citep{luo2016, lu2017} --- we optimize textures directly in 3D scene space, ensuring physical consistency across frames as objects scale during looming. We evaluate models under both adversarial attacks and common video perturbations (e.g., flicker, fog), and pair each robustness measurement with computational cost in FLOPs per inference to characterize robustness relative to efficiency.

The main contributions of our work are:
\begin{itemize}
    \item A differentiable 3D rendering pipeline for a binary collision detection task (approaching objects that hit or miss the observer) in which adversarial attacks optimize object textures for differentiable motion-perception models.  
   
    \item A direct comparison of gradient-based adversarial attacks and the reverse-phi illusion on the same connectome-constrained, task-optimized model of the fly visual system (FlyNet). 
    
    \item Using \emph{counterfactual restoration}, we find textures equally far from the training distribution that recover classification accuracy and position and velocity readouts, linking failure to motion-code disruption, not unfamiliar textures.

    \item We identify the computation both perturbations exploit: Adversarial textures and reverse-phi both invert the Hassenstein--Reichardt correlation between radially neighboring inputs, and optimizing this statistic alone is sufficient to induce or reverse classification failure.
    
    \item We benchmark robustness--efficiency trade-offs across architectures with different inductive biases (recurrence, global signals, continuous-time dynamics) under adversarial attacks and common video perturbations.
            
    \item We find that FlyNet and sCTRNN have higher robustness per order of magnitude of compute than every other architecture on common video corruptions of the collision-detection task.     
\end{itemize}

\section{Background}

\paragraph{Visual motion detection in the fly and FlyNet}

In the \textit{Drosophila} optic lobe (Supp. Fig.~\ref{fig_supp:flyvis}a), visual motion detection is decomposed into parallel ON and OFF pathways, mediated by the T4 and T5 neurons, respectively \citep{Maisak2013}. These neurons act as local moving-edge detectors, with T4 and T5 selective for ON and OFF edges, each with subtypes (a, b, c, d) selective for the four cardinal directions of motion, forming a distributed basis for local motion encoding \citep{haag2017}. These local directional signals are subsequently integrated by Lobula Plate-Lobula Columnar neurons, such as LPLC2, which are selectively tuned to radial expansion patterns associated with looming, and then project to giant fiber descending neurons driving escape behaviors such as jumping \citep{Borst2015, Klapoetke2017, Ache2019Looming}. This hierarchical integration encodes collision-relevant variables ---including radial expansion, object position, and velocity--- establishing a shared representational substrate from which multiple task-relevant variables can be decoded. 
There has been extensive computational work applying structural constraints from the fruit fly on neural networks \citep{Tschopp2018-od, mano_predicting_2021, morra2023multifunctionality, Lappalainen2024}. Here we use FlyNet from \citep{Lappalainen2024}, which provides a computational instantiation of the early motion detection circuit (excluding LPLC2, and giant fiber neurons), modeling neurons as passive-point units governed by
\begin{equation}\label{eq:dynamics}
    \tau_i \dot{h}_i = -h_i + \sum_j w_{ij} f(h_j) + h^{\text{rest}}_{i} + l_i,
\end{equation}
where synaptic weights $w_{ij}$ are constrained by connectomic measurements while neuronal and synaptic parameters are optimized for motion detection using an optic flow dataset (see App.~\ref{app:flynet}).


\paragraph{Visual motion illusion in the fly: Reverse-phi} Biological visual systems estimate motion accurately under natural conditions, but illusory stimuli that deviate from natural scene statistics elicit systematic perceptual errors, reflecting adaptations of neural circuitry to natural environments~\citep{Eagleman2001-mh}. A canonical example is the reverse-phi illusion, when the contrast of a moving pattern is inverted between successive frames ---bright regions become dark and vice-versa--- the perceived direction of motion reverses, in humans \citep{anstis1970, anstis1975, Agrochao2020}, macaques \citep{livingstone2001}, and flies~\citep{tuthill2011neural, Agrochao2020}. Flies, when presented with a reverse-phi stimulus, turn their body in the direction opposite to the physical motion, a reflex that reflects the inversion of direction-selective T4/T5 signals~\citep{clark2011structure, tuthill2011neural, Leonhardt2017, Gruntman2021-io}.

\paragraph{Adversarial attacks in optical flow} Neural networks are vulnerable to small, targeted perturbations that induce misclassification \citep{szegedy2014}. Such perturbations can be designed under worst-case gradient-based attacks (PGD \citep{Madry2018}, CW \citep{carliniwagner2017}), gradient-free methods (NES \citep{wierstra2008nes}), and common video perturbations such as noise, flicker, and fog \citep{Hendrycks2019-ez, oei2026robustspring}. Such adversarial examples often transfer across models, revealing shared vulnerabilities \citep{papernot2016}. In optical flow specifically, patch-based adversarial attacks can severely corrupt motion estimation \citep{ranjan2019, schrodi2022}, and global norm-constrained perturbations show that state-of-the-art networks are particularly vulnerable \citep{schmalfuss2022}. However, pixel-space and patch-based perturbations often lose effectiveness under minor geometric transformations \citep{luo2016, lu2017} as they do not capture how perturbations evolve under coherent object motion. Optimizing textures over 3D objects addresses this in still images \citep{athalye2018}. In optical flow, adversarial particles rendered consistently in 3D degrade methods that are robust to per-pixel attacks \citep{Schmalfuss2023Distracting}, but adversarial textures on moving objects remain unexplored.

\section{Methods}

\begin{figure}[t!]
    \centering
    \includegraphics[width=\textwidth, trim=0cm 2.85cm 0.5cm 0cm, clip]{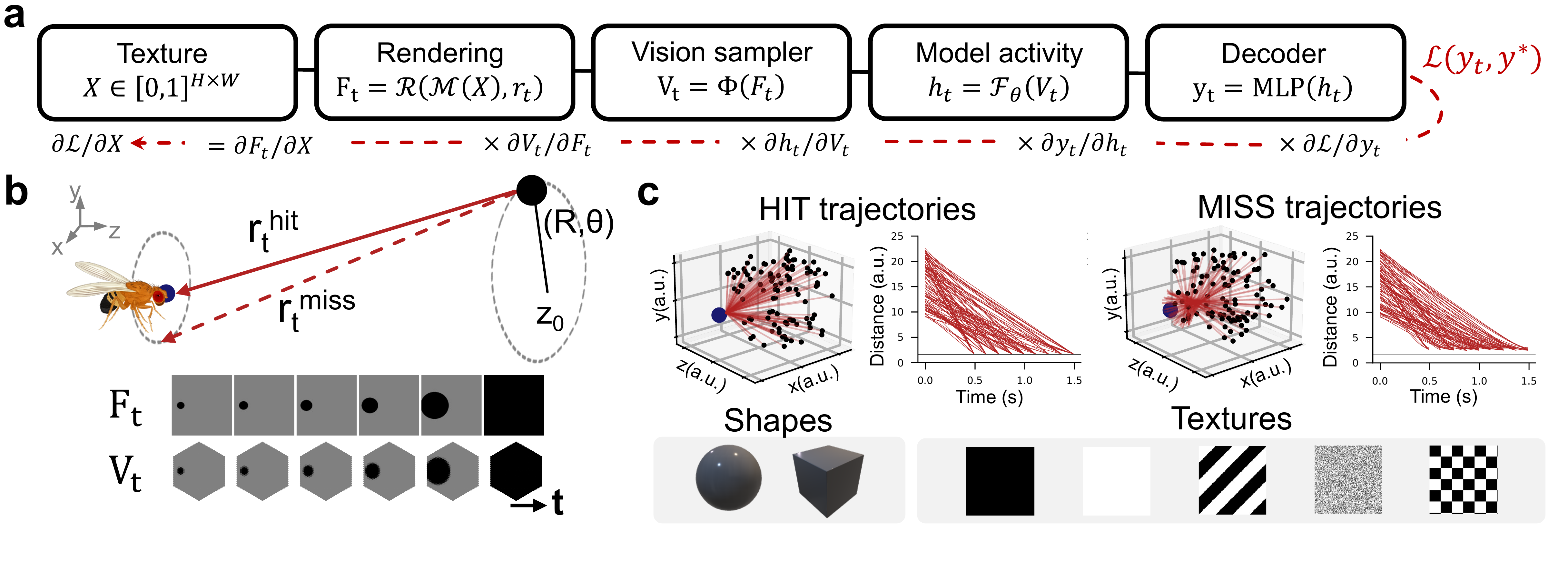}
    \caption{
    \textbf{Fully differentiable 3D geometric framework for motion-based collision detection.}
    \textbf{(a)} Processing differentiable pipeline. \textbf{(b)} Geometric setup of looming stimuli, parameterized by polar coordinates $(R, \theta, z_0)$. Bottom: example rendering in the image frame $F_t$ and corresponding fly visual input $V_t$. \textbf{(c)} Generated dataset: example trajectories, shapes and textures. Examples of hit and miss trajectories in 3D space and their temporal distance profiles.}    
    \label{fig:framework}
\end{figure}

\subsection{A differentiable 3D rendering pipeline for a looming motion perception task}

We assess motion perception as a binary decision task in a visual looming paradigm, where an approaching object must be classified as \textit{hit} or \textit{miss} \citep{Zhou22_looming}. This task is a canonical, behaviorally grounded paradigm in \textit{Drosophila} \citep{muijres2014, Ache2019Looming}. To generate stimuli, we develop a differentiable 3D rendering pipeline implemented in \texttt{pytorch3d} \citep{ravi2020pytorch3d}, which provides full control over scene geometry and ensures that physical perturbations evolve consistently with the underlying dynamics across time (Figure~\ref{fig:framework}). We render grayscale sequences at $416 \times 416$ pixels and $100$ Hz, and, when required, map them onto a hexagonal visual sampler before feeding them to each model. The resulting model activity is then used by a Multi-Layer Perceptron (MLP) decoder to produce the final decision (Figure~\ref{fig:framework}a). All stimuli are defined in the reference frame of the fly-observer, modeled as a point-particle camera fixed at the origin. We initialize objects at distance $z_0$ from the origin, with positions parameterized by polar coordinates $(R,\theta)$ in the plane orthogonal to the $z$-axis, enabling isotropic sampling of trajectories across the visual field (Figure~\ref{fig:framework}b). Each trajectory consists of a linear translation in 3D space over duration $T_{\text{traj}}$, directed either toward the origin (\textit{hit}) or toward a nearby offset endpoint (\textit{miss}). This construction yields controlled deviations from collision while preserving comparable motion statistics across conditions. The resulting trajectories span angular sizes from ${\sim}5$--$110^\circ$, with peak angular expansion speeds of up to $1{,}420^\circ/\mathrm{s}$, consistent with the looming-escape literature: $5$--$90^\circ$ \citep{Ache2019Looming} and $650$--$2{,}100^\circ/\mathrm{s}$ \citep{muijres2014}. We define each stimulus by its trajectory parameters $\left(R,\theta,z_0,T_{\text{traj}}\right)$ and object appearance parameters, including shape $\{\text{Sphere}, \text{Cube}\}$ and texture $X \in \{\text{Black}, \text{White}, \text{Checkerboard}, \text{Wavy}, \text{Random}\}$, with $X \in \mathbb{R}^{128 \times 128}$ (Figure~\ref{fig:framework}c). Random textures are generated from pixelwise uniform noise, offset by a uniformly sampled constant to keep the local mean different from background. 

\subsection{Architectures}

\paragraph{Continuous-time recurrent neural networks} FlyNet is derived from the connectome-constrained recurrent model of the \textit{Drosophila} visual system \citep{Lappalainen2024}, with $45{,}669$ neurons and 1.51M anatomically constrained edges, trained on Sintel optical flow \citep{Butler_2012} with only 734 free parameters.
We edited the weights of two pretrained model instances to create a hybrid model, exhibiting accurate direction selectivity for all T4 and T5 subtypes (App.~\ref{app:flynet}, Supp. Fig.~\ref{fig_supp:flyvis}). As a control, we evaluate CTRNN, a FlyNet variant with the same dynamics, input, and cell-type connectivity, but without single-cell connectivity, synapse counts and synapse signs from the connectome, trained under the same Sintel protocol. It has 7.51M edges, connecting cells of the 604 connected cell-type pairs through isotropic convolutional connectivity over a hexagonal kernel of 19 columns (five columns across), yielding 11,593 free parameters \citep{Lappalainen2024}. As an edge-count-matched control, we additionally evaluate a sparsity-matched CTRNN variant (sCTRNN). Starting from the trained CTRNN, we untied the shared filter weights, pruned per-edge by the mean absolute injected current, down to FlyNet's 1.51M connections, and fine-tuned on Sintel. It recovered optical-flow performance to an EPE 2\% below FlyNet's on held-out Sintel samples. 
All models here are evaluated using instantaneous T4/T5 responses (5,768 units) as decoder input (see App.~\ref{app:flynet},~\ref{app:ctrnn}, and~\ref{app:sctrnn} for details). Their dynamics follow Eq. \ref{eq:dynamics}, integrated using Euler with $dt=0.01$. They receive as input the luminance of the visual stimulus sequence, sampled to a regular hexagonal lattice of 721 columns.

\paragraph{Optical flow networks} We evaluate four feedforward CNN baselines following a FlowNet-inspired architecture \citep{Dosovitskiy2015flownet} (${\sim}414$k parameters), pretrained on Sintel and using the same hexagonal visual sampling as FlyNet. The vanilla CNN (vCNN) processes 2 consecutive frames using local receptive fields, while its global variant (gCNN) incorporates channel-wise global pooling into the second hidden layer (Supp. Fig.~\ref{fig_supp:cnn_architectures}) to probe the role of \textbf{global integration}. We also train two vCNN variants using 5 and 10 input frames, respectively, while keeping the remaining architecture unchanged; to assess the effect of \textbf{temporal context}. For all CNN variants, features are extracted from the final hidden layer (7,688 activations). As a higher-performance baseline, we evaluate RAFT~\citep{RAFT2020}, pretrained and fine-tuned on Sintel (${\sim}5.3\text{M}$ parameters). We extract pooled features from the final hidden representation (6,802 activations) and vary recurrent iterations (1, 6, 12) to probe the role of \textbf{recurrence}. We also evaluate RAFT-small (RAFTsm, ${\sim}1\text{M}$ parameters) as a lower-parameter baseline. These models receive the original $416\times416$ luminance frames replicated across three channels. To control for differences in input representation and training regime, we additionally trained RAFT hex-matched variants exclusively on Sintel using the same hexagonally sampled input as FlyNet and the CNN baselines (hexRAFT; Supp. Fig.~\ref{fig:raft_hex_sampler}).

\subsection{Optimizations and attacks}
\paragraph{Decoders} Task-specific MLPs are trained on frozen network representations on ${\sim}600{,}000$ samples from 6,000 trials, and tested on ${\sim}40{,}000$ held-out samples from 400 trials. Each trial contributes 50--150 timepoints (100\,Hz, durations 0.5--1.5\,s), treated as independent samples. We train a binary hit/miss classifier ($512$ hidden units, $\tanh$) and a continuous position/velocity regressor ($128-16$ hidden units, ReLU) (App.~\ref{app:decoder_training}, Supp. Fig.~\ref{fig_supp:decoder_training},~\ref{fig_supp:decoder_test_accuracies_a},~\ref{fig_supp:decoder_test_accuracies_b},~\ref{fig_supp:motion_decoders}). Both trained with Adam~\citep{Kingma2015-mk} ($\text{lr}=10^{-3}$, batch size $1024$).
Our decoder architecture is inspired by how the fly visual system is organized for collision-detection: The fly's LPLC2 neurons detect radial expansion patterns and, by nonlinearly integrating dendritic inputs from T4 and T5 neurons (spatially), trigger the animal's escape response \citep{Klapoetke2017}. Since this mechanism integrates spatial patterns, we used a nonlinear, spatial-only decoder architecture to abstract LPLC2 computation.
%
%
\paragraph{Adversarial optimization} We optimize texture $X$ on hit trials to suppress the decoder's predicted probability of a collision, $p(\text{hit})$. As in targeted adversarial optimization~\citep{szegedy2014, carliniwagner2017}, the loss combines the cross-entropy $\mathcal{L}_{CE}$ toward the target label with an $L_2$ penalty on the distance to a baseline texture $X_{\text{init}}$:
\begin{equation}\label{eq:opt_illusion}
    \mathcal{L}(X) = \mathcal{L}_{CE}(X, y_{\text{target}}=0) + 
    \lambda_{\text{init}}\| X-X_{\text{init}} \|_2.
\end{equation}
The resulting texture, $X_{\text{illusion}}$, minimizes the above loss.

\paragraph{Counterfactual restoration} We find the nearest texture to $X_{\text{illusion}}$ that recovers correct classification, akin to optimization-based counterfactual explanations~\citep{wachter2017counterfactual}, minimizing:
\begin{equation}\label{eq:opt_reverse}
    \mathcal{L}(X) = \mathcal{L}_{CE}(X, y_{\text{target}}=1) + 
    \lambda_{\text{init}}\| X-X_{\text{illusion}} \|_2.
\end{equation}
Both stages use Adam~\citep{Kingma2015-mk} with learning rate $0.01$ and $\lambda_{\text{init}}=0.02$.

\paragraph{Hassenstein--Reichardt looming detector (HR)} Given $C_i(t)=V_t^i-\bar{V}(t)\;$ the input at hexal $i$ minus the mean over all hexals, and $\mathcal{P}$ the set of radially neighboring hexal pairs $(i, j)$, with $i$ the inner and $j$ the outer one, we compute the HR correlation \citep{Hassenstein1956-dn} for a given delay $\tau$ as (Supp. Fig.~\ref{fig_supp:hrc_cartoon})
\begin{equation}\label{eq:hrc}
  g(\tau) \equiv g(X, \tau) \;=\; \big\langle\,
  C_i(t-\tau, X)\, C_j(t, X) \;-\; C_j(t-\tau, X)\, C_i(t, X)
  \,\big\rangle_{t,\,(i,j)\in\mathcal{P}} \,,
\end{equation}
where $g(\tau)>0$ indicates looming, $g(\tau)<0$ contraction, and $g(\tau)=0$ static or non-directional input. Note that the computation depends only on the input, making it agnostic to the model. The optimization process finds the texture that minimizes or maximizes Eq.~\ref{eq:hrc}.

\paragraph{Adversarial attacks} All attacks optimize the logit of the object texture, on 100 time-steps hit trajectories with Black initial texture. For PGD \citep{Madry2018}, we use 100 steps with step size $\varepsilon/10$, an adaptive step-size scheme \citep{croce2020autoattack}, 2 random restarts, and 15 independent attacks per condition; resulting textures are applied directly to other models to assess transfer attacks. For CW \citep{carliniwagner2017}, we minimize a distance plus misclassification objective. For NES \citep{wierstra2008nes}, we estimate gradients via Gaussian perturbations. See App.~\ref{app:adv_attacks} for details.

\paragraph{Video perturbations} We test six common video perturbations \citep{Hendrycks2019-ez, oei2026robustspring}: additive white noise (AWN) on the texture, AWN on the frames, shot (Poisson) noise, frame flickering, motion jitter, and contrast reduction (fog), with noise power matched across architectures (see App.~\ref{app:video_perturbations} for details; Supp. Fig.~\ref{fig_supp:environmental_stressors_examples}). AWN, shot noise, flickering, and motion jitter are averaged over 10 trajectories with 10 noise realizations each, fog over 20 trajectories.

\paragraph{Robustness $\mathcal{R}$} For each attack, we sweep a parameter defining the magnitude of the perturbation and define robustness as the area under the accuracy curve, which we call $\mathcal{R}$ \citep{kunhardt2021auc}.

\section{Results}

\subsection{Adversarial failures arise from disrupted motion representations}
\begin{figure}[t]
    \centering
    \includegraphics[width=\textwidth]{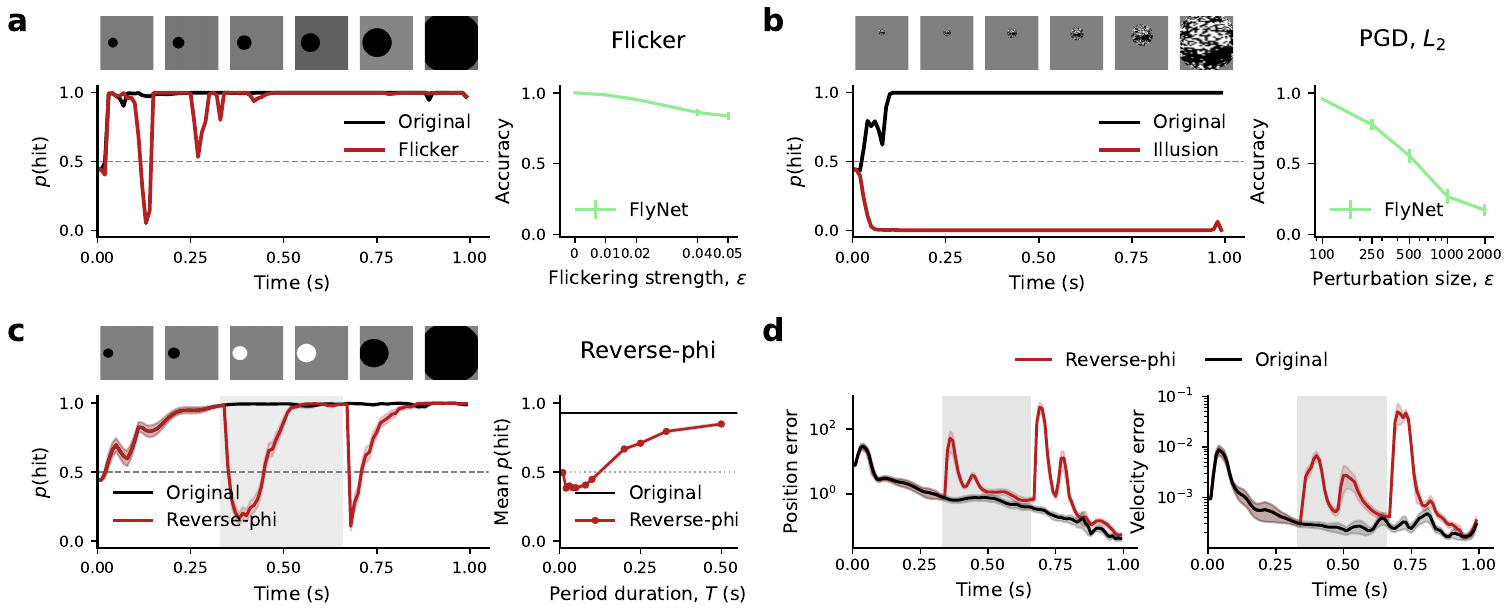}
    \caption{
    \textbf{FlyNet is robust to flicker but sensitive to adversarial and reverse-phi attacks.}
    \textbf{(a)}~$p(\text{hit})$ under flicker ($\varepsilon=0.05$) and accuracy vs.\ flicker strength (10 trajectories $\times$ 10 noise samples).
    \textbf{(b)}~$p(\text{hit})$ under an $L_2$-PGD adversarial texture ($\varepsilon=2000$) and accuracy vs.\ perturbation size (15 trajectories $\times$ 2 restarts).
    \textbf{(c)}~$p(\text{hit})$ under reverse-phi ($T=0.330 \, s$) and average $p(\text{hit})$ vs.\ contrast-inversion period (50 trajectories).
    \textbf{(d)}~Position and velocity decoding errors under reverse-phi vs.\ original texture ($T=0.330 \, s$; 50 trajectories). Gray regions indicate contrast-inversion windows; error bars and shaded curves denote standard error of the mean (SEM).
    }

    \label{fig:attacks_fly}
\end{figure}

FlyNet maintains hit/miss classification accuracy under common video corruptions -- additive white noise, shot noise, flicker, motion jitter, and contrast reduction (Fig.~\ref{fig:attacks_fly}a; Supp.~Fig.~\ref{fig_supp:environmental_stressors_curves},~\ref{fig_supp:environmental_stressors_examples}) -- but accuracy plummets under optimized adversarial attacks targeting the looming object's texture (Fig.~\ref{fig:attacks_fly}b; Supp.~Fig.~\ref{fig_supp:adversarial_curves},~\ref{fig_supp:adversarial_NES},~\ref{fig_supp:cw_curves},~\ref{fig_supp:flynet_adversarial_examples}). We asked what the adversarial perturbations are exploiting. Do they disrupt the motion representations that FlyNet was trained to compute, or do they push the input outside the training texture distribution?

\paragraph{Reverse-phi disrupts motion representations with familiar textures} Reverse-phi is a known illusion in real flies that disrupts motion representations. It inverts the contrast between successive frames, evoking illusory reversed motion through inversion of T4/T5 direction-selective signals \citep{clark2011structure, tuthill2011neural, Leonhardt2017}. Using familiar textures, we constructed reverse-phi looming stimuli by inverting the object's contrast every $T$ seconds. Under this stimulus, $p(\text{hit})$ drops sharply at each contrast inversion and recovers between them (Fig.~\ref{fig:attacks_fly}c), mirroring the optomotor inversion seen in real flies \citep{tuthill2011neural}.
Position and velocity decoded from FlyNet activity show the same dependence: decoding errors spike at each inversion (Fig.~\ref{fig:attacks_fly}d) and depend on $T$. Short $T$ gives the network too little time to recover, long $T$ returns both $p(\text{hit})$ and decoding errors toward baseline (Supp.~Fig.~\ref{fig_supp:reverse_phi_pos_vel_vs_T}). Reverse-phi establishes that motion representations can be disrupted without any input-texture distribution shift.

\paragraph{Counterfactual restoration rules out texture novelty as the cause of adversarial failure}

\begin{figure}[t]
    \centering
    \includegraphics[width=\textwidth]{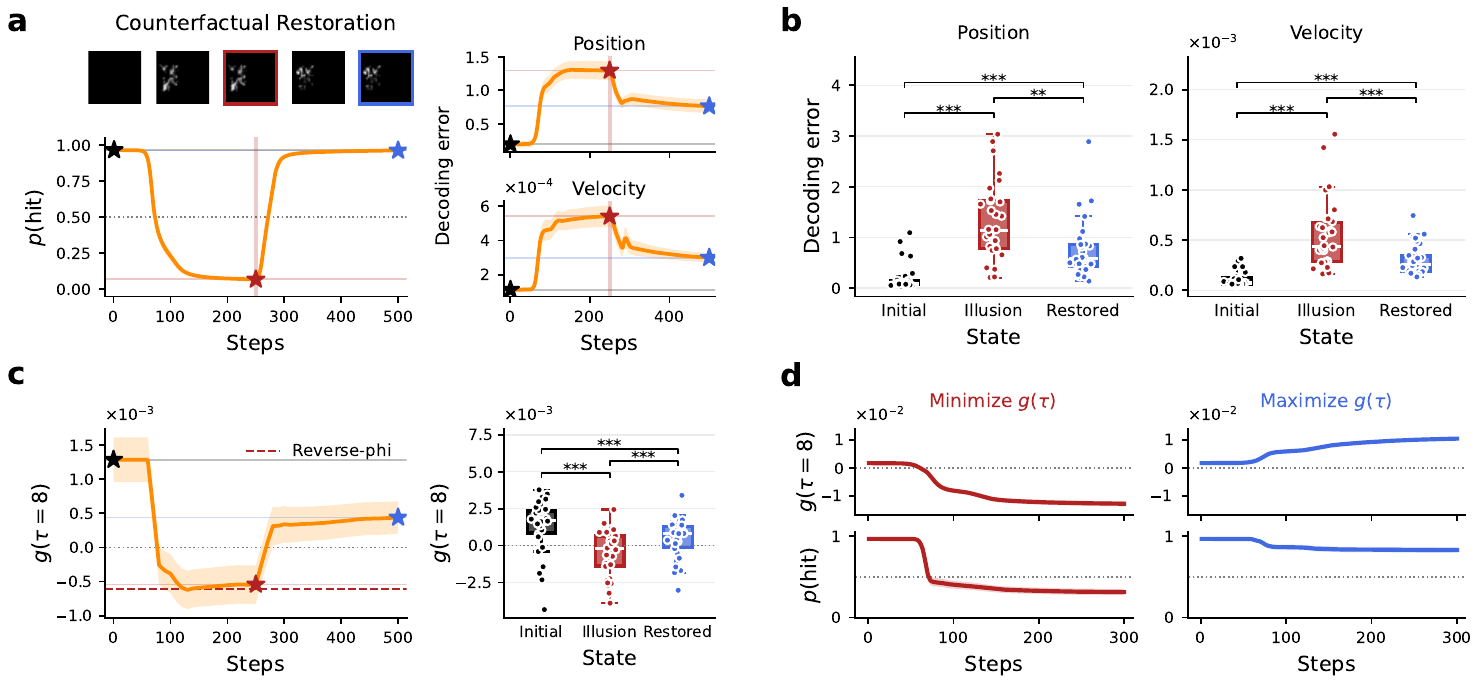}
    \caption{
    \textbf{A static adversarial texture and a dynamic illusion exploit the same low-level motion computation.}
    \textbf{(a)}~$p(\text{hit})$ and position and velocity decoding errors across adversarial optimization and counterfactual restoration (30 trajectories). Stars mark initial (black), illusion (red), and restored (blue) textures; texture evolution is shown above.
    \textbf{(b)}~Position and velocity decoding errors at the three counterfactual restoration checkpoints.
    \textbf{(c)}~HR correlator $g(\tau=8)$ (Eq.~\ref{eq:hrc}) across counterfactual restoration steps (left), and at the three checkpoints (right). 
    \textbf{(d)}~Minimization (left) and maximization (right) of $g(\tau=8)$ (top) and corresponding $p(\text{hit})$ (bottom) across optimization steps.
    Shaded area represents SEM; asterisks indicate significance from paired $t$-test ( **: $p<0.01$, ***: $p<0.001$).
    }
    \label{fig:disruptions_fly}
\end{figure}

To test whether adversarial failure arises from the same kind of within-distribution disruption, we optimized a texture to suppress $p(\text{hit})$ on a hit trajectory (Eq.~\ref{eq:opt_illusion}), then optimized the nearest texture that recovers correct classification (Eq.~\ref{eq:opt_reverse}). The restored texture remains as far from the training distribution as the illusion texture throughout optimization (Supp. Fig.~\ref{fig_supp:distance_textures_counterfactual}). If unfamiliar textures were the cause of failure, restored textures should disrupt accuracy as strongly, which they don't. Position and velocity readouts recover alongside $p(\text{hit})$ (Fig.~\ref{fig:disruptions_fly}a,b; Supp. Fig.~\ref{fig_supp:decoding_over_steps_counterfactual}) consistent with the adversarial restoration recovering the same motion code that reverse-phi disrupts. Single T4/T5 traces show that the illusion texture shifts activity in the same direction as reverse-phi during the inversion phase, consistent with both perturbations engaging T4/T5 motion detectors in the same direction (Supp. Fig.~\ref{fig_supp:tracesndisruption}). To quantify this effect at the population level, we compared responses across all 5,768 T4/T5 neurons in a 330-ms window aligned to reverse-phi contrast inversions ($n=30$ trajectories). Activity induced by the adversarial illusion was the closest to reverse-phi's, compared to both the original and the restored conditions (all metrics $p<0.005$; Supp. Fig.~\ref{fig_supp:rsa}; see App.~\ref{app:rsa}), suggesting that adversarial textures and reverse-phi may exploit similar computational vulnerabilities.

\paragraph{Inverting a second-order motion statistic suffices to induce failure} We hypothesized that both perturbations act on the same second-order motion statistic: The delayed correlation between neighboring inputs that Hassenstein--Reichardt correlators read out \citep{Hassenstein1956-dn}, to which T4/T5 direction selectivity is sensitive \citep{clark2011structure}, and which reverse-phi is known to invert \citep{tuthill2011neural}. To test this, at each step of counterfactual restoration, we computed the HR correlator (Eq.~\ref{eq:hrc}). We found that it closely tracks the motion decoding errors (Fig.~\ref{fig:disruptions_fly}c; Supp. Fig.~\ref{fig_supp:comparison_phit_gtau}a). Supporting our hypothesis, when the network is maximally fooled, HR reaches values close to the ones from reverse-phi (Fig.~\ref{fig:disruptions_fly}c, Supp. Fig.~\ref{fig_supp:hrc_cartoon}b). To test whether disrupting this statistic is sufficient for failure, we separately minimized and maximized $g(\tau= 8 \,\mathrm{frames} = 80\,\mathrm{ms})$~(Eq.~\ref{eq:hrc}), without access to the network. Under minimization $p(\text{hit})$ drops to $\sim 0.3$, whereas under maximization it stays above $0.8$ (Fig.~\ref{fig:disruptions_fly}d; Supp. Fig.~\ref{fig_supp:comparison_phit_gtau}b,~\ref{fig_supp:hr_correlator_failrestore}). Manipulating this statistic alone is thus sufficient to induce the failure, while increasing it leaves classification intact, suggesting that the adversarial attacks indeed exploit this canonical, well-studied motion statistic.

Adversarial failures in FlyNet are therefore not explained by unfamiliar texture statistics. Adversarial textures and reverse-phi invert the same second-order correlation and shift T4/T5 population activity in the same direction, attributing both perturbations to a common vulnerability in the motion representations of FlyNet.

\subsection{Robustness profiles are shaped by structural characteristics}

\begin{figure}[t!]
    \centering
    \includegraphics[width=\textwidth]{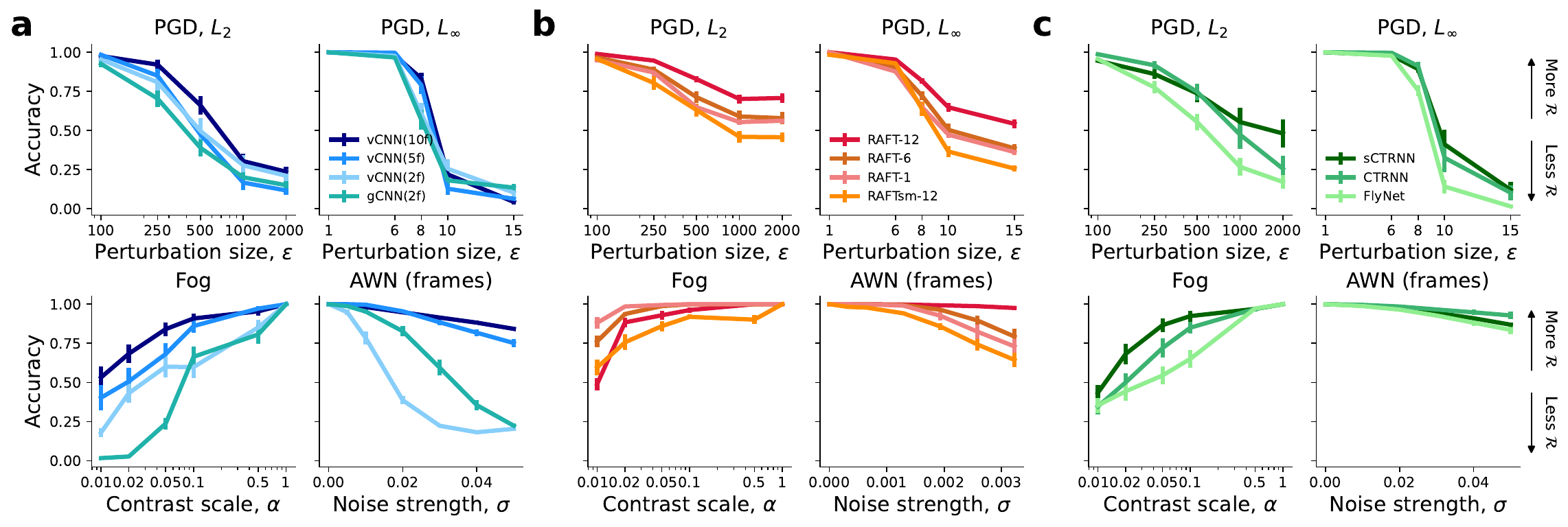}
    \caption{\textbf{Global integration, recurrence, temporal context, and connectivity shape robustness under adversarial and common video perturbations.}
    \textbf{(a)} Effect of global signal integration (vCNN vs.\ gCNN) and temporal context (vCNN(2 frames) vs.\ vCNN(5 frames) vs.\ vCNN(10 frames)).
    \textbf{(b)} Effect of recurrence depth (RAFT family).
    \textbf{(c)} Effect of connectivity (FlyNet vs.\ sCTRNN vs.\ CTRNN). Top row: accuracy under PGD attacks 
    ($L_2$ and $L_\infty$). Bottom row: two representative common video perturbations (AWN on frames and fog). Error bars represent SEM.
    }
    \label{fig:ingredients_robustness}
\end{figure}

FlyNet's architecture is predominantly local and feedforward, reflecting the columnar organization of the fly optic lobe \citep{Fischbach1989-cr}. 
This is at odds with what has been typically used in ANNs to provide robustness: recurrence and global integration \citep{Kar2019-da, Spoerer2020-em, Geirhos2018-zd, Naseer2021-hd, Yan2019-sf, Dapello2020-hv}.
We asked whether these ingredients are required for robustness in general, and which best account for FlyNet's robustness profile. 
\paragraph{Global integration helps conditionally}
We compared two feedforward CNN baselines that differ only in global pooling: a vanilla CNN (vCNN) with purely local receptive fields, and a variant with an additional channel-wise global pooling pathway that injects a spatially averaged scene representation into the network (gCNN; Supp.~Fig.~\ref{fig_supp:cnn_architectures}). gCNN's pooled global signal helps when it carries perturbation-relevant information that local features lack -- additive noise on frames, flicker and high-photon shot noise are all spatially distributed and gCNN beats vCNN under each (Fig.~\ref{fig:ingredients_robustness}a; Supp. Fig.~\ref{fig_supp:environmental_stressors_curves}). It hurts when the global signal carries little perturbation-relevant information but dominates a weakened local signal -- fog, low-photon shot noise, and adversarial textures all reduce local contrast against a still-present scene background, and gCNN underperforms vCNN under each (Fig.~\ref{fig:ingredients_robustness}a; Supp. Fig.~\ref{fig_supp:adversarial_curves},~\ref{fig_supp:adversarial_NES}; Supp. Fig.~\ref{fig_supp:environmental_stressors_curves}). Global pooling therefore buys robustness conditionally: it does not provide unconditional robustness, and it introduces specific failure modes where the pooled scene average can overwhelm weak local signals.

\paragraph{Recurrence improves robustness to white-box attacks}
The RAFT model \citep{RAFT2020} allows applying an arbitrary number of recurrent refinement iterations. Increasing recurrence depth (1, 6, and 12) in RAFT improved robustness in white-box attack conditions: CW attack success decreased monotonically, and accuracy under PGD increased substantially from RAFT-6 to RAFT-12, with RAFT-1 and RAFT-6 similar. RAFT-12 was also the most robust under additive noise on frames, shot noise, and flicker (Fig.~\ref{fig:ingredients_robustness}b; Supp. Fig.~\ref{fig_supp:adversarial_curves},~\ref{fig_supp:cw_curves},~\ref{fig_supp:environmental_stressors_curves}; Table~\ref{table:robustness_full}). Robustness was unchanged under the black-box NES and transfer attacks and under noise on texture, and decreased under motion jitter and fog, where RAFT-12 was the most vulnerable of the family (Supp. Fig.~\ref{fig_supp:adversarial_NES},~\ref{fig_supp:transfer_matrices},~\ref{fig_supp:environmental_stressors_curves}; Table~\ref{table:robustness_full}). In hexRAFT, recurrence had little or inconsistent effect. Recurrent refinement is therefore a candidate for robustness to white-box attacks in artificial networks. FlyNet is also recurrent, but it updates its state once per frame rather than refining its state over repeated iterations on the same input, so the RAFT result cannot directly carry over to FlyNet.

\paragraph{Temporal integration supports robustness to common video corruptions} FlyNet, CTRNN, and sCTRNN integrate their input through continuous-time dynamics (Eq.~\ref{eq:dynamics}), which low-pass filter perturbations that change from frame to frame and provide more temporal context than the two-frame input of CNN and RAFT baselines. 
All three continuous-time models are highly robust to motion jitter, flicker and additive noise on frames, above most baselines (Fig.~\ref{fig:ingredients_robustness}c; Supp. Fig.~\ref{fig_supp:environmental_stressors_curves}; Table~\ref{table:robustness_full}), consistent with this low-pass filtering.
To manipulate temporal context directly in a network without continuous-time dynamics, we retrained vCNN with 5 and 10 input frames instead of 2. Robustness was higher with 5 or 10 frames than with 2 under most perturbation types, and under all common video corruptions (Fig.~\ref{fig:ingredients_robustness}a; Supp. Fig.~\ref{fig_supp:environmental_stressors_curves}; Table~\ref{table:robustness_full}), showing that temporal context improves robustness to common video corruptions in a feedforward network.

\paragraph{No single connectivity property orders robustness among continuous-time models}
The three continuous-time models share dynamics and input but differ in connectivity.
FlyNet's connectome-derived structure is the most local: The influence-weighted mean distance between perturbed input columns and responding neurons (the projective counterpart \citep{lehky1988_receptive_field} of the effective receptive field \citep{luo2016_receptive_field_cnns}; see App.~\ref{app:d_hat}) is $\bar{d}\approx1.5$ hexagonal columns, compared to $2.2$ for sCTRNN and $2.8$ for CTRNN ($2.7$ for vCNN and $4.3$ for gCNN).
CTRNN, with ${\sim}5\times$ more edges, is more robust than FlyNet under most perturbations tested, although the gap is small on common video corruptions (Fig.~\ref{fig:ingredients_robustness}c;  Supp. Fig.~\ref{fig_supp:adversarial_curves},~\ref{fig_supp:adversarial_NES},~\ref{fig_supp:transfer_matrices},~\ref{fig_supp:cw_curves},~\ref{fig_supp:environmental_stressors_curves}; Table~\ref{table:robustness_full}). 
Pruning CTRNN to FlyNet's edge count and fine-tuning it (sCTRNN) increased robustness under PGD $L_2$ and fog beyond both CTRNN and FlyNet, and matched CTRNN under PGD $L_\infty$ (Fig.~\ref{fig:ingredients_robustness}c), but reduced it below CTRNN under $L_\infty$ transfer attacks, and below both CTRNN and FlyNet under flicker (Supp. Fig.~\ref{fig_supp:adversarial_NES},~\ref{fig_supp:transfer_matrices},~\ref{fig_supp:environmental_stressors_curves}; Table~\ref{table:robustness_full}). The ordering therefore depends on the perturbation, and neither the number of free parameters (Table~\ref{table:parameter_count}), edge count nor locality explain it consistently across these conditions. 
Under every adversarial attack, FlyNet, the most local model with the fewest free parameters, is the least robust of the three.

Together, these results show that robustness to common video corruptions is consistent with temporal integration and increases when temporal context is added to vCNN, whereas no single tested connectivity property explains the differences among the continuous-time models.


 \subsection{Robustness-efficiency trade-offs}

 \begin{figure}[t!]
    \centering
    \includegraphics[width=\textwidth]{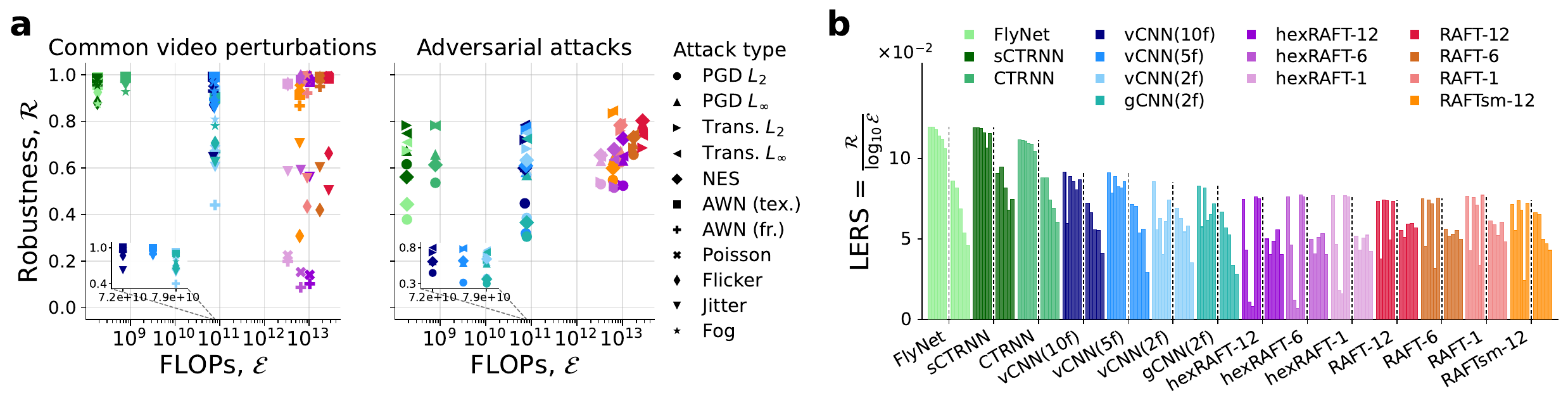}
    \caption{\textbf{Robustness-efficiency frontier across perturbation regimes.} \textbf{(a)} Robustness $\mathcal{R}$ vs. computational cost $\mathcal{E}$, measured in FLOPs, for each network and perturbation type. Left: common video perturbations, including texture noise, frame noise, Poisson noise, flicker, jitter and fog. Right: adversarial attacks, including PGD, transfer and NES attacks. Inset: zoom for the CNN family.
    \textbf{(b)} LERS computed for all networks and perturbation types. Common video perturbations
    (left group): AWN texture, motion jitter, shot noise, AWN frames, 
    flicker, fog. Adversarial attacks (right group): transfer $L_\infty$, 
    transfer $L_2$, PGD $L_\infty$, NES, PGD $L_2$.}
    \label{fig:robustness_efficiency}
\end{figure}
Architectures span five orders of magnitude in compute cost:
FlyNet runs at $0{.}186$ GFLOPs per inference, RAFT-12 at $27{,}402$ GFLOPs, with CNN baselines and CTRNN in between (Fig.~\ref{fig:robustness_efficiency}a; Table~\ref{table:robustness_full}). Compute does not buy proportional robustness. Within the RAFT family, robustness rises with cost under white-box attacks; but vCNN and gCNN sit at comparable FLOPs and differ substantially in robustness, and on common video corruptions the fly-inspired models achieve high robustness at a fraction of the cost of any other architecture tested. How much robustness does each architecture buy per unit of compute?

\paragraph{Log-Efficiency Robustness Score}
We measure the robustness-efficiency trade-off as 
\begin{equation}
    \text{LERS} = \frac{\mathcal{R}}{\log_{10} \mathcal{E}},
\end{equation}
where $\mathcal{E}$ is FLOPs per inference. The logarithmic normalization reflects diminishing returns in robustness from added compute \citep{bartoldson2024adversarial}, and is conservative -- it penalizes compute less than linear normalizations would, so any trade-off advantage observed for low-compute architectures would only grow under stricter normalization. LERS represents robustness per order of magnitude of compute (Fig.~\ref{fig:robustness_efficiency}b, Table~\ref{table:robustness_full}).

\paragraph{Sparsity places FlyNet in a favorable LERS regime on common video corruptions}
We computed LERS for each model and perturbation type (Fig.~\ref{fig:robustness_efficiency}b, Table~\ref{table:robustness_full}). Within the CNN and RAFT families, compute differs by less than 4\% in $\log_{10}\mathcal{E}$, so LERS follows the robustness rankings reported above. Across families, compute spans five orders of magnitude, and on all six common video corruptions the three continuous-time models have the highest LERS of all models.
The one within-group comparison in which LERS departs from robustness is FlyNet versus CTRNN. On common video corruptions, robustness is high for all three continuous-time models ($\mathcal{R}\geq0.87$), and CTRNN's ${\sim}4\times$ higher compute raises its $\log_{10}\mathcal{E}$ by ${\sim}7.5\%$, more than its robustness lead over FlyNet; FlyNet therefore has higher LERS than CTRNN on common corruptions, although CTRNN is more robust. FlyNet and sCTRNN, with nearly identical compute, are ordered by $\mathcal{R}$ alone and reach comparable LERS. Under PGD $L_2$ and NES, CTRNN's robustness lead exceeds the compute difference, and both CTRNNs rank above FlyNet. sCTRNN is more robust than FlyNet at slightly lower compute and has the highest LERS of all models under PGD and transfer attacks.
%
%
%
Together, these results show that FlyNet's low edge count, and hence low compute, gives it higher LERS than CTRNN and all baselines other than sCTRNN on common video corruptions, despite CTRNN's higher raw robustness. sCTRNN, matched to FlyNet's edge count, reaches comparable LERS without FlyNet's synapse-level wiring.

\section{Discussion}

Robust and efficient sensory processing is a hallmark of biological neural circuits, yet the architectural principles that support this combination remain poorly understood. Here, we asked whether synaptic wiring derived from the connectome contributes to either property beyond what standard artificial architectures and task training achieve. To answer this, we built a differentiable 3D rendering pipeline for looming-based collision detection and benchmarked a connectome-constrained, task-optimized model of the fly visual system (FlyNet) against unconstrained CNN, globally pooled CNN, CTRNN, and RAFT-based baselines under adversarial attacks and common video perturbations. Recurrence (in RAFT), global context, and temporal context each support robustness under different perturbation regimes. FlyNet's connectome constraint, however, does not seem to contribute to robustness --- CTRNN and sCTRNN, which lack detailed connectome constraints, are more robust than FlyNet under almost every perturbation tested. What the connectome seems to contribute is efficiency through sparsity: FlyNet matches CTRNN closely on common video corruptions while using a fraction of the synapses and compute, yielding higher LERS than any architecture other than sCTRNN. Our comparison to sCTRNN suggests that this efficiency advantage is not unique to the connectome's specific wiring. While FlyNet is robust to the tested common video corruptions it remains sensitive to optimized adversarial attacks (gradient-based PGD, gradient-free NES) indicating that biological structure alone does not imply general adversarial robustness. The fly itself is susceptible to a structured stimulus: Reverse-phi inverts the contrast of moving patterns and produces illusory motion in real flies --- a perceptual error that reflects a specific computational vulnerability of the T4/T5 motion-detection circuit. FlyNet inherits this vulnerability, and our analysis shows that adversarial optimization discovers stimuli that exploit the same failure mode: Both invert the Hassenstein--Reichardt correlation between neighboring inputs, and manipulating that second-order correlation statistic alone suffices to induce the failure. 

\paragraph{Limitations}

In the fly, downstream looming-selective neurons, such as LPLC2 and LC4, project to escape pathways \citep{Klapoetke2017, Ache2019Looming} that are not present in FlyNet. Beyond connectivity, biological circuits compute under additional constraints such as neuromodulation, synaptic plasticity, synaptic stochasticity \citep{greco2026}, and spiking --- none of which FlyNet models. Some asymmetries complicate cross-architecture comparison.
First, decoder input dimensionality differs across architectures, which could confound cross-architecture robustness comparisons: a higher-dimensional readout may itself contribute to measured robustness rather than the backbone alone, leaving this as an axis for future work.
Second, sCTRNN and FlyNet share edge count and compute but beyond wiring differ by more than three orders of magnitude in trainable parameters (1,513,361 vs 734; Table~\ref{table:parameter_count}; Supp. Fig.~\ref{fig:fig_supp_degree}), as well as in training pipeline (sCTRNN is fine-tuned), so differences between them cannot be attributed to synapse-level connectome constraints alone.
Third, our robustness and efficiency results hold for the binary looming task, while generalization to other visual tasks, such as object tracking or navigation, and under more naturalistic scenes, remains untested and left as future work.
Fourth, each architecture is represented by a single trained instance, so error bars reflect stimulus and attack variability, not variability across trained networks.
Finally, our compute estimate uses FLOPs, which enables consistent cross-architecture comparison but does not measure the metabolic cost of biological computation \citep{Laughlin1998, Attwell2001-ia, Sengupta2010, Deistler2022-cz}.

\paragraph{Conclusion and future direction}

In general, adversarial vulnerability appears to be a consequence of specialized computational strategies:
Networks become robust to the perturbations they were trained on and vulnerable to perturbations designed against their architecture. Adversarial training is the field's canonical answer, raising the question whether the connectome's wiring already sits closer to an adversarially trained solution than an artificial neural network trained from scratch. 
Differentiable connectome-constrained models also turn adversarial attacks into a tool that can interface with experiments in biology. Future connectome-constrained models that incorporate downstream circuits and greater biological realism will generate more faithful predictions of adversarial stimuli for experimental tests.
Because the correlation-based motion motif is shared with mammalian motion vision \citep{clark2014_flies_humans_corr, Borst2015}, the approach could extend to larger connectome models as a route to candidate illusions for mammalian circuits. 
Our work suggests adversarial analysis as a tool for probing the limits of biological computation, while also informing the design of specialized neural networks that achieve favorable robustness--efficiency trade-offs. Immediate future work will focus on isolating the role of the connectome in robustness and address the generalization of these findings across more tasks.

\newpage
\begin{ack} \label{sec:ack}
We thank the organizers and participants of the Cajal NeuroAI summer school 2025 at the Champalimaud Foundation, Lisbon, where this project was initiated. We also thank Srini Turaga, Mason McGill, and Eyal Gruntman for the discussions and helpful comments. This work was supported by MCIN/AEI/10.13039/501100011033 and FSE+ (grants PID2022-143311NB-I00, PRE2022-101492, and RYC2022-037595-I); by the Spanish State Research Agency, through the Severo Ochoa and María de Maeztu Program for Centers and Units of Excellence in R\&D (CEX2020-001084-M);  by a Newcastle University Academic Track (NUAcT) Fellowship and PhD studentship; by the Lister Institute Prize Fellowship; by the Air Force Office of Scientific Research (AFOSR, FA9550-23-1-0533); by the Leverhulme Research Leadership Award; by the Shanahan Foundation Fellowship at the Interface of Data and Neuroscience (Shanahan Family Foundation and Allen Institute); by the Howard Hughes Medical Institute (HHMI), Janelia Research Campus (we thank Srini Turaga for compute support); by the European Union (ERC, DeepCoMechTome, 101089288 to Jakob Macke). Views and opinions expressed are however those of the author(s) only and do not necessarily reflect those of the European Union or the European Research Council. Neither the European Union nor the granting authority can be held responsible for them; and by the ACCESS program (project: CIS261112, allocation: Anvil, Purdue University \citep{anvilgpu}), supported by NSF grants \#2138259, \#2138286, \#2138307, \#2137603, and \#2138296. We thank CERCA Programme/Generalitat de Catalunya for institutional support. The code and networks associated with this work will be available at \url{https://github.com/alexgarciaduran/flynet-robustness}, \url{https://huggingface.co/alejandrorg00/flynet-robustness}, and the flyvis fork \url{https://github.com/lappalainenj/flyvis} (release \texttt{neurips2026}).
\end{ack}

\section*{Author contributions}
Conceptualization, methodology: A.G.-D., A.R.-G. and J.K.L.
Software and investigation: A.G.-D., A.R.-G. and J.K.L.
Analysis: A.G.-D., A.R.-G. and J.K.L.
Writing: A.G.-D., A.R.-G. and J.K.L.
Writing (review and editing): M.M.-M., A.H., S.R.
Supervision: J.K.L.
Funding: M.M.-M., A.H., S.R. and J.K.L.

\section*{Competing interests}
The authors declare no competing interests.

{\small
\bibliographystyle{unsrtnat}
\bibliography{references_merged}
}

\newpage
\appendix
\setcounter{figure}{0}
\section{Additional information}
\subsection{Architectures}
\subsubsection{FlyNet} \label{app:flynet}
FlyNet \citep{Lappalainen2024} is a differentiable neural network whose architecture is derived from a synapse-level connectome reconstruction of the \textit{Drosophila} optic lobe, covering 65 cell types. The network topology, which neurons are connected and with what anatomical synapse counts, is fixed by connectomic measurements and not modified during training. What is learned are the unknown biophysical parameters: the single-neuron time constants $\tau_i$, resting potentials $h_i^{rest}$, and synaptic strength parameters that scale the fixed anatomical weights, yielding only 734 free parameters in total. These parameters are optimized using standard deep learning techniques to minimize an optical flow prediction loss on the Sintel dataset \citep{Butler_2012}. Crucially, no electrophysiological data are used during training: the model recovers single-neuron response properties consistent with three decades of experimental measurements as an emergent consequence of fitting the motion task under the connectome constraint. FlyNet has 45,669 neurons and 1.51M synaptic connections, and is governed by the continuous-time dynamics in Eq. \ref{eq:dynamics}, where $f$ is a ReLU nonlinearity applied to pre-synaptic voltages, and $l_i$ represents the luminance input to photoreceptors (for all other cell types, $l_i = 0$).

\begin{figure}[ht]
    \centering
    \includegraphics[width=\linewidth]{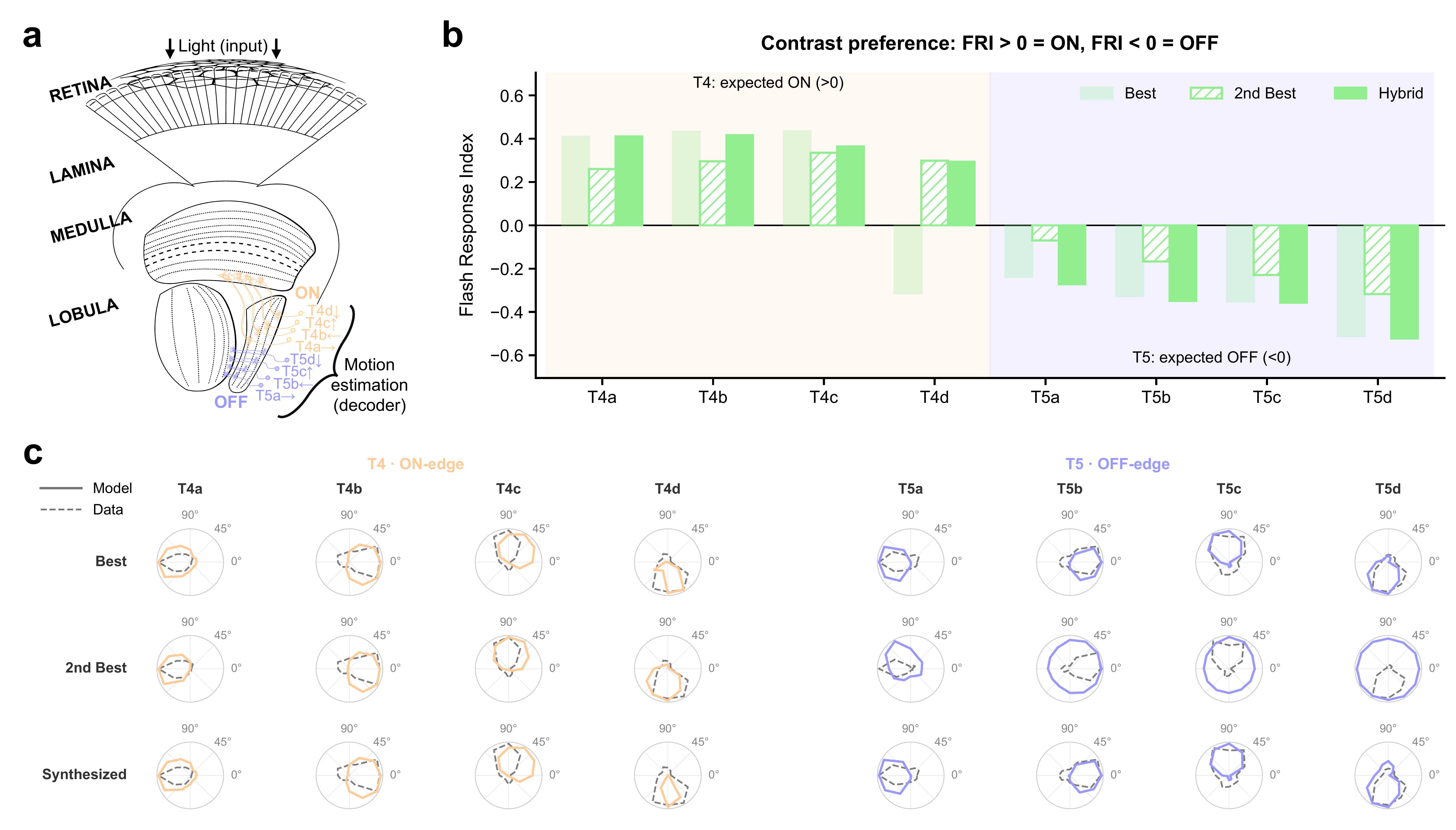}
    \caption{\textbf{FlyNet visual circuit.} \textbf{(a)} Schematic of the fly optic lobe~\citep{borst2020}, from retina to lobula, highlighting ON and OFF motion-selective T4/T5 pathways used for motion estimation by the decoder. \textbf{(b)} Contrast preference of the T4 and T5 subtypes, measured by the flash response index (FRI; ON $>0$, OFF $<0$), for the best-loss model, the second-best model (T4d donor), and the hybrid model. The parameter substitution reverses T4d from OFF to the expected ON preference and preserves the contrast preference of all other subtypes. \textbf{(c)} Directional tuning of T4 (ON-edge) and T5 (OFF-edge) neurons across the three models (solid lines), compared with experimental data (dashed lines).}
    \label{fig_supp:flyvis}
\end{figure}

We specifically decode T4 and T5 motion pathways (Supp. Fig.~\ref{fig_supp:flyvis}). The original work reports neural tuning averaged over clusters of 50 equivalently configured and stochastically trained model instances, and individual instances can deviate from known tuning properties. However, no single pretrained model exhibits correct direction selectivity for all T4 and T5 subtypes simultaneously. In the model with the best Sintel validation loss, T4d prefers OFF instead of ON contrast and is only weakly direction selective, whereas all other T4 and T5 subtypes have the expected contrast preference and preferred direction. We therefore constructed a hybrid model: starting from the best-loss model, we replaced the resting potential and time constant of T4d, and the synaptic strengths of all ten connection types onto T4d, with those of the second-best model, whose T4d prefers ON contrast. After this substitution, all T4 and T5 subtypes have the expected contrast preference and preferred direction, and T4d becomes strongly direction selective (Supp. Fig.~\ref{fig_supp:flyvis}b,c).

\subsubsection{CTRNN} \label{app:ctrnn}
CTRNN follows the same continuous-time dynamics as FlyNet (Eq. \ref{eq:dynamics}), receives the same hexagonal input, but its weights $w_{ij}$ are not constrained by the connectome. CTRNN learns weights of free signs for all connected cell-type pairs over a support of a 19-column hexagonal neighborhood (five columns across), increasing edge count from 1.51M to 7.51M and increasing the number of free parameters from $734$ to $11{,}593$. CTRNN thus serves as an ablation of the synapse-level connectome prior while holding the dynamical framework constant. CTRNN is the best-Sintel-validation network from Extended Data Fig. 2, DMN[11,593] in \citep{Lappalainen2024}.

\subsubsection{sCTRNN}\label{app:sctrnn}

sCTRNN is obtained from the trained CTRNN by untying the shared filter weights and pruning per edge by the mean absolute injected current, down to FlyNet's 1.51M connections, followed by fine-tuning on Sintel. sCTRNN matches FlyNet's edge count of 1,513,231.

\begin{figure}[ht]
\centering
\includegraphics[width=\linewidth]{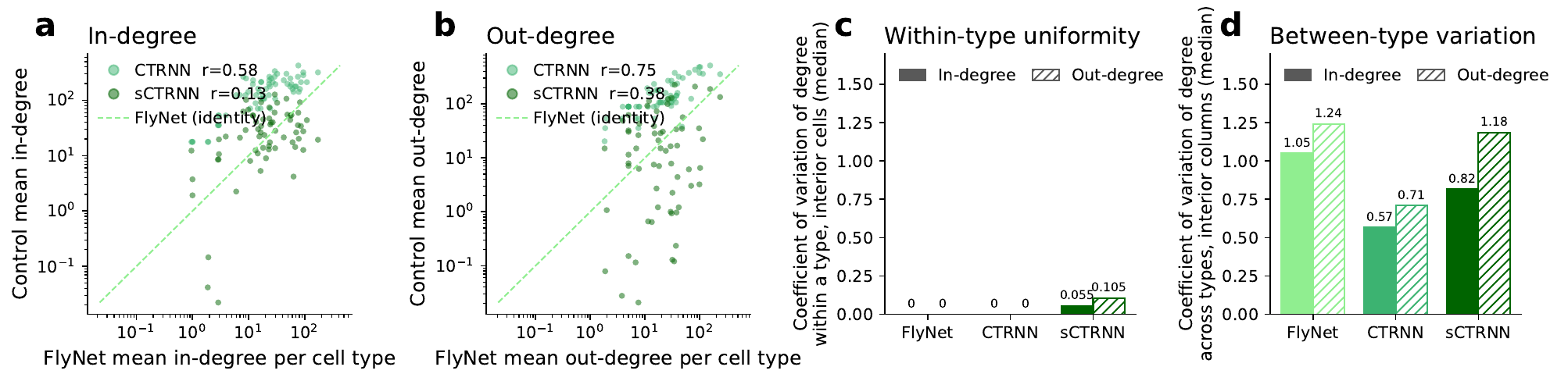}
\caption{\textbf{Comparison of in-degree and out-degree statistics between FlyNet, CTRNN, and sCTRNN.}
\textbf{(a)}~Mean in-degree per cell type in CTRNN and sCTRNN against FlyNet (each point one of 65 cell types; dashed line: identity). $r$: Pearson correlation with FlyNet's per-type mean degree.
\textbf{(b)}~Same for out-degree.
\textbf{(c)}~Coefficient of variation of degree within a cell type's interior cells (interior meaning 10 of 15 columns across to remove boundary effects), median across types: FlyNet, CTRNN, and sCTRNN share one filter support by construction. sCTRNN's differs by a few edges, because of the remainder from pruning the isotropic convolutional filter support from 7.51M to 1.51M edges.
\textbf{(d)}~Coefficient of variation of degree across cell types within each interior column, median across columns. CTRNN's dense filters flatten degree differences between types.}
\label{fig:fig_supp_degree}
\end{figure}

Supp. Fig.~\ref{fig:fig_supp_degree}a,b characterize remaining differences in mean degree per cell type between sCTRNN, CTRNN, and FlyNet. CTRNN and sCTRNN degree distributions differ from FlyNet's ($r=0.13$ in, $0.38$ out). Supp. Fig.~\ref{fig:fig_supp_degree}c, d characterize within and across cell-type in-degree and out-degree variation in each network.  

\subsubsection{CNNs}
\begin{figure}[ht]
    \centering
    \includegraphics[width=0.95\linewidth]{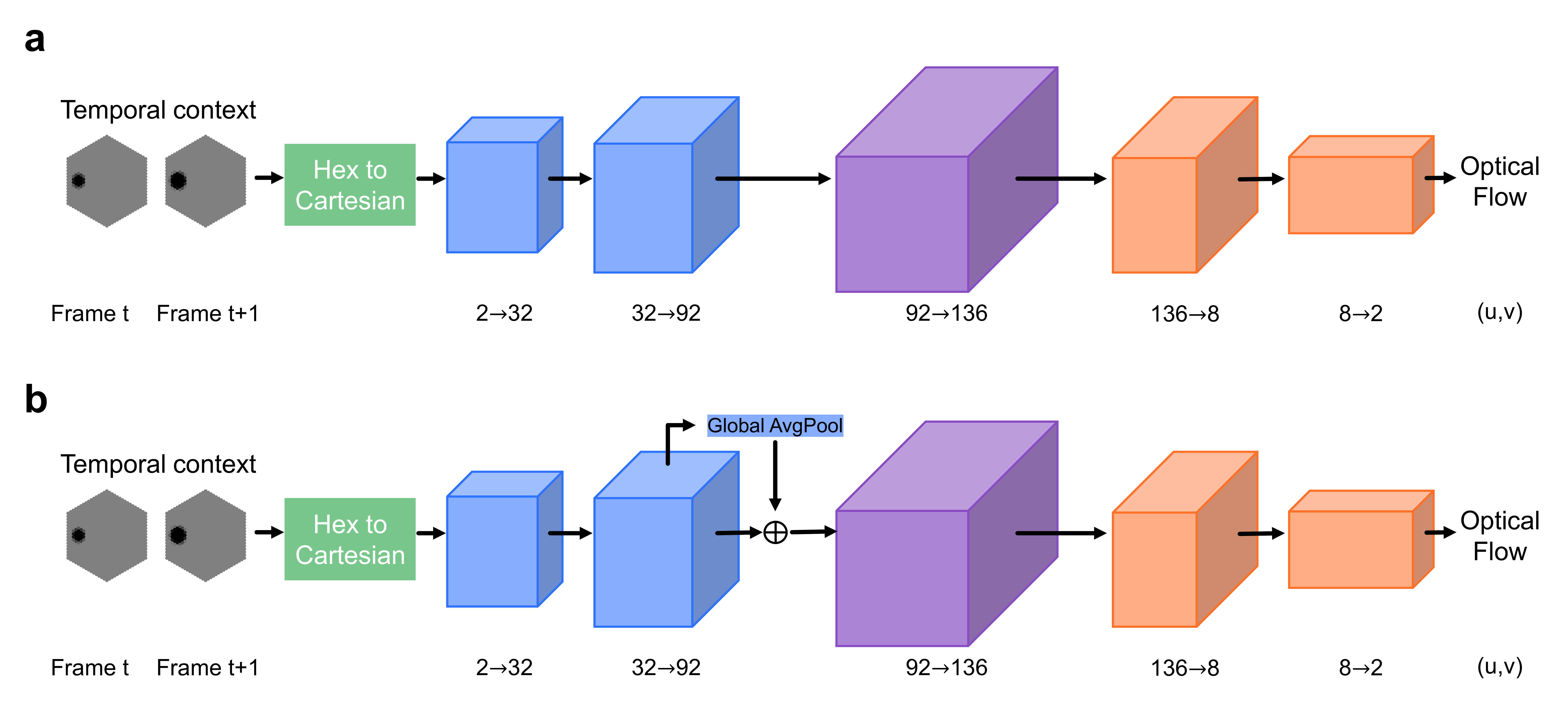}
    \caption{vCNN \textbf{(a)} and gCNN \textbf{(b)} architectures.}
    \label{fig_supp:cnn_architectures}
\end{figure}

We considered unconstrained CNN baselines trained on Sintel. The vanilla CNN consists of five layers with 32, 92, 136, 8, and 2 channels, respectively. All layers use a kernel size of 5, except for the last layer, which uses a kernel size of 1. Each layer consists of a convolution, batch normalization, and an exponential linear unit activation, except for the final layer, which applies only a convolution. The global pooling network uses the same architecture, but includes a global average pooling operation between the second and third layers, providing the network with access to global spatial context (Supp. Fig.~\ref{fig_supp:cnn_architectures}). Among our CNNs, only vCNN(2f) is one of five networks from Extended Data Fig. 2, vanilla CNN in \citep{Lappalainen2024}.

\subsubsection{RAFT}
We use RAFT as a high-performance optical flow baseline, pretrained on standard benchmarks, and fine-tuned on Sintel. We extract feature maps from the final hidden representation prior to flow output. Given its size $128 \times H/8 \times W/8$ (we use $H=W=416$), we apply 2D average pooling (kernel size 7, stride 7), resulting in $6,802$ features for decoding.

\begin{figure}[ht]
    \centering
    \includegraphics[width=0.75\linewidth]{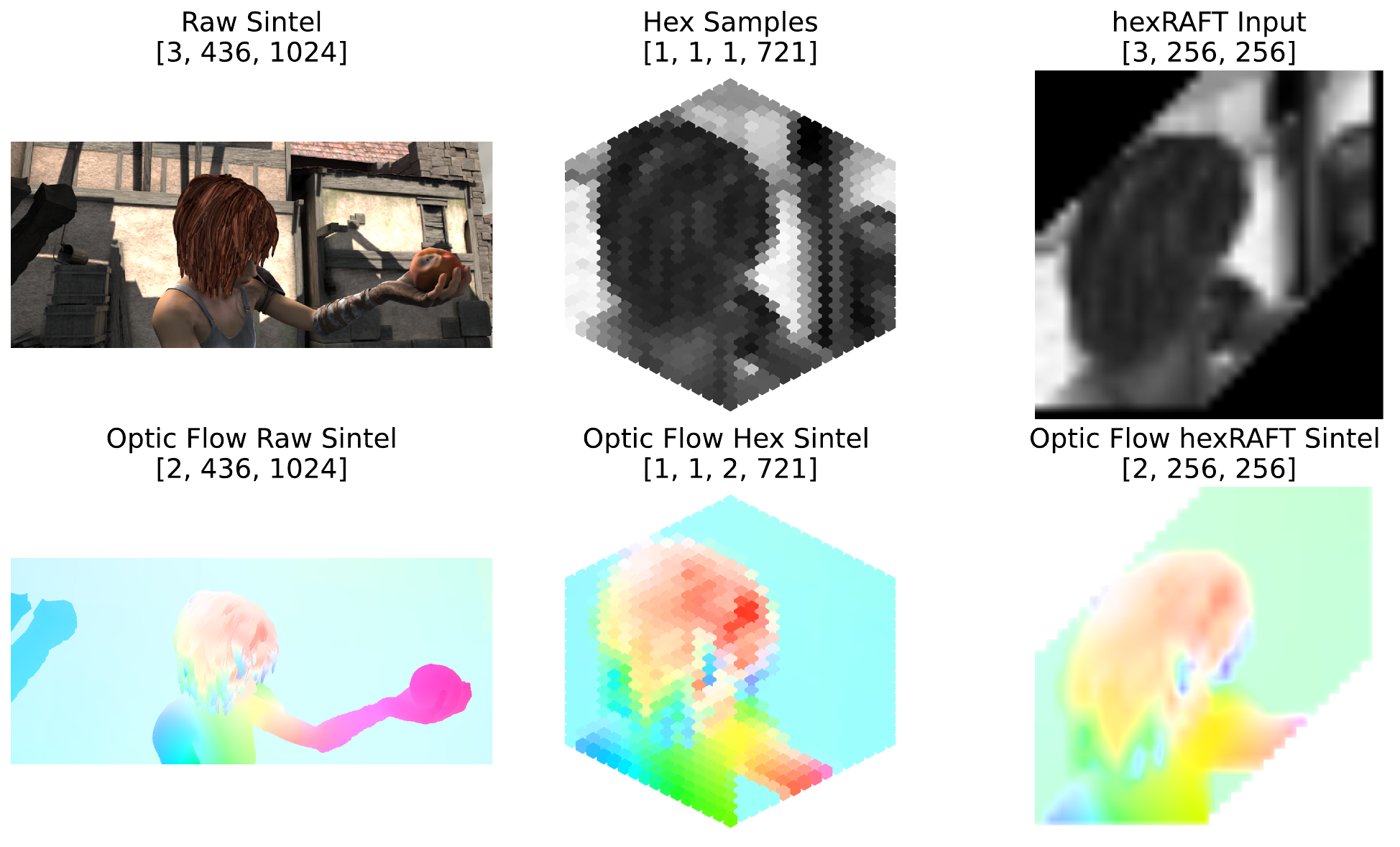}
    \caption{\textbf{Hexagonal sampling pipeline for hexRAFT Sintel.} Sintel RGB frames (top) and ground-truth optical flow (bottom) are shown in their original resolution (left), after hexagonal sampling with 721 receptors (middle), and after conversion to the $256\times256$ Cartesian representation used by RAFT (right). This matches the hexagonal sampling geometry of FlyNet, as do the other CTRNN and CNN benchmark families.}
    \label{fig:raft_hex_sampler}
\end{figure}

To control for differences in visual sampling, we additionally trained a FlyNet-matched RAFT model (hexRAFT) using the same hexagonal sampling geometry as FlyNet (Supp. Fig.~\ref{fig:raft_hex_sampler}). Sintel frames were converted to luminance and sampled through an array of 721 hexagonally arranged photoreceptors, each performing spatial averaging over a $13\times13$ pixel receptive field. The resulting samples were mapped onto a $31\times31$ Cartesian grid, zero-padded to $32\times32$, and bilinearly upsampled to $256\times256$ to produce RAFT-compatible inputs. The same sampling procedure was applied to the ground-truth optical flow, without rescaling flow magnitudes, and a validity mask excluded unsampled grid locations from the training loss. We trained the model on the Clean and Final passes of Sintel using the same scene-disjoint training split, for $50,000$ iterations with a batch size of 8, using AdamW ($\mathrm{lr}=10^{-4}$, weight decay $=10^{-5}$) and $12$ recurrent update iterations.

\subsection{Simulation}
We used \texttt{pytorch3d} \citep{ravi2020pytorch3d} to render the stimulus videos. The camera is fixed at the origin $\vec{r}_c = (0, 0, 0)$. Objects are initialized at distance $z_0$ along the $z$-axis, with positions in the plane parameterized by polar coordinates $(R, \theta)$: $\vec{r}_0 = (R\cos\theta,\, R\sin\theta,\, z_0)$. Each trajectory is a linear translation of duration $T_{\text{traj}}$ from $\vec{r}_0$ to a final point $\vec{r}_f$. For hit trials, $\vec{r}_f = (0, 0, z_\text{offset})$, where $z_\text{offset}$ is a small displacement along $\hat{k}=(0,0,1)$ to prevent the object from intersecting the camera. For miss trials, the endpoint is shifted to the antipodal point on the circle: $\vec{r}_f = (-\alpha x_0 + \kappa_1,\, -\alpha y_0 + \kappa_2,\, z_\text{offset})$, with $\alpha = 0.25$, $\kappa_1 = \cos(\pi + \operatorname{atan2}(y_0, x_0))$, and $\kappa_2 = \sin(\pi + \operatorname{atan2}(y_0, x_0))$. Trajectory parameters were sampled uniformly: $\theta \in [0, 2\pi)$, $R \in [3, 6]$, $z_0 \in [8, 22]$, and $T_{\text{traj}}$ took one of 9 equally spaced values in $[0.5, 1.5]$ s.

\subsection{Functional locality}\label{app:d_hat}
We ask how far a network carries the effect of a single input column --- the projective counterpart \citep{lehky1988_receptive_field} of the effective receptive field \citep{luo2016_receptive_field_cnns}. Let $i$ index the units of a network, each assigned the coordinate $\mathbf{p}_i$ (in units of the nearest-neighbor column spacing) of the hexal it sits above. Perturbing hexal $j$ by $0.25$ above a uniform input of $0.5$ and time-averaging each unit's response, we define the influence of $j$ on unit $i$, $\delta^{(j)}_i$, as the absolute change from the unperturbed run. Repeating this for every input column $j$, locality is the influence-weighted mean distance $\bar{d} = \sum_{i,j} d^{(j)}_i \delta^{(j)}_i \big/ \sum_{i,j} \delta^{(j)}_i$, with $d^{(j)}_i = \lVert \mathbf{p}_i - \mathbf{p}_j \rVert$ the distance from the perturbed hexal. It is the center of mass of the perturbation's effect: small when a column only moves its neighbors, larger as it spreads across the lattice.

\subsection{Robustness quantification}
For each attack type, we sweep a parameter defining the magnitude of the perturbation and define robustness as the area under the accuracy curve $\mathcal{R}$ \citep{kunhardt2021auc}:
\begin{equation*}
\begin{aligned}
\mathcal{R} &= \frac{1}{x_{max}-x_{min}}\int_{x_{\min}}^{x_{\max}} \mathrm{Acc}(x)\, dx, 
\quad
\mathrm{Acc}(x) = \frac{1}{NT} \sum_{i=1}^{N} \sum_{t=1}^{T} \mathbf{1}\!\left[\hat{y}_{i,t}(x) = y_{i,t}\right].
\end{aligned}
\end{equation*}
For CW attacks \citep{carliniwagner2017}, we quantify robustness as the minimum perturbation magnitude across values of $c$ satisfying an attack success criterion of at least 80\% misclassified frames:
\begin{equation*}
    \delta^*_r = \min_{c:\text{success}_{r,c} \geq 0.8} \delta_{r,c} \, ,
\end{equation*}
where $\delta_{r,c}$ is the perturbation magnitude ($L_2$ or $L_\infty$) for run $r$ and trade-off coefficient $c$.

\subsection{FLOPs quantification}
We estimated inference FLOPs of all networks using the PyTorch profiler (\texttt{torch.profiler}). FLOPs are computed for 100-frame random videos.

\subsection{Compute}
We ran experiments on Google Cloud Vertex AI using NVIDIA L4 and A100 GPUs, and on Purdue University’s Anvil cluster using NVIDIA A100 GPUs, as computationally required.
\newpage
\subsection{Decoding}\label{app:decoder_training}
Both decoders are trained with Adam~\citep{Kingma2015-mk} ($\text{lr}{=}10^{-3}$, batch size 1,024). The binary hit/miss classifier uses 512 hidden units, $\tanh$ activation, cross-entropy loss, $L_2$ regularization $\alpha{=}0.1$, max 50 epochs, and early stopping with patience 10. The continuous position/velocity regressor uses 128--16 hidden units, ReLU activation, MSE loss, max 200 epochs, and early stopping with patience 50 and min improvement $10^{-5}$.

\subsubsection{Hit/Miss}
\begin{figure}[ht]
    \centering
    \includegraphics[width=\linewidth]{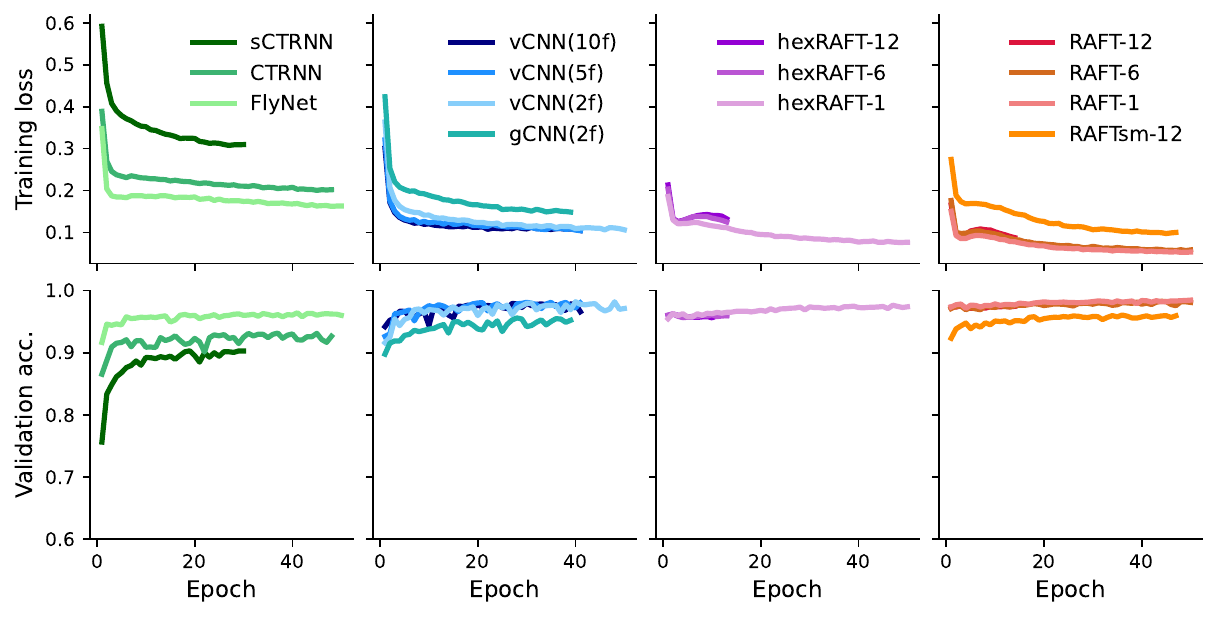}
    \caption{\textbf{Training dynamics of the hit/miss decoder across model architectures.} Training loss (top) and validation accuracy (bottom) as a function of training epoch for FlyNet, CTRNN, sCTRNN, CNN-based models, and Cartesian and hexagonally sampled RAFT variants. Models are grouped by architecture across columns, with each curve representing a different model configuration.}
    \label{fig_supp:decoder_training}
\end{figure}

\begin{figure}[p]
    \centering
    \includegraphics[width=0.95\linewidth]{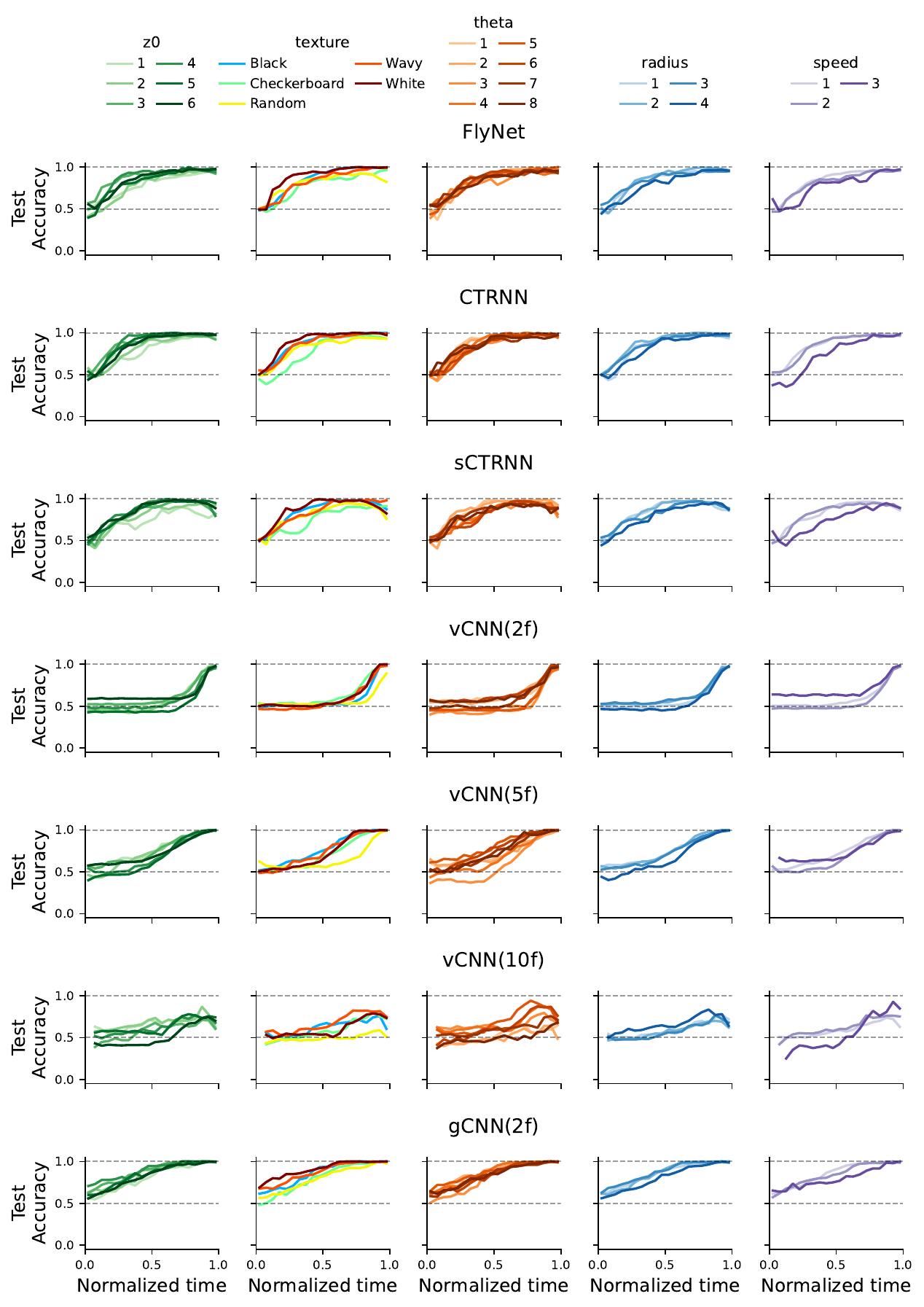}
    \caption{Decoder test accuracy over time for all CTRNN and CNN networks, with normalized time (dividing by its total duration), conditioning on parameters of the simulation. Numbers in the legend correspond to ascending bins of the uniformly sampled parameters.}
    \label{fig_supp:decoder_test_accuracies_a}
\end{figure}

\begin{figure}[p]
    \centering
    \includegraphics[width=0.95\linewidth]{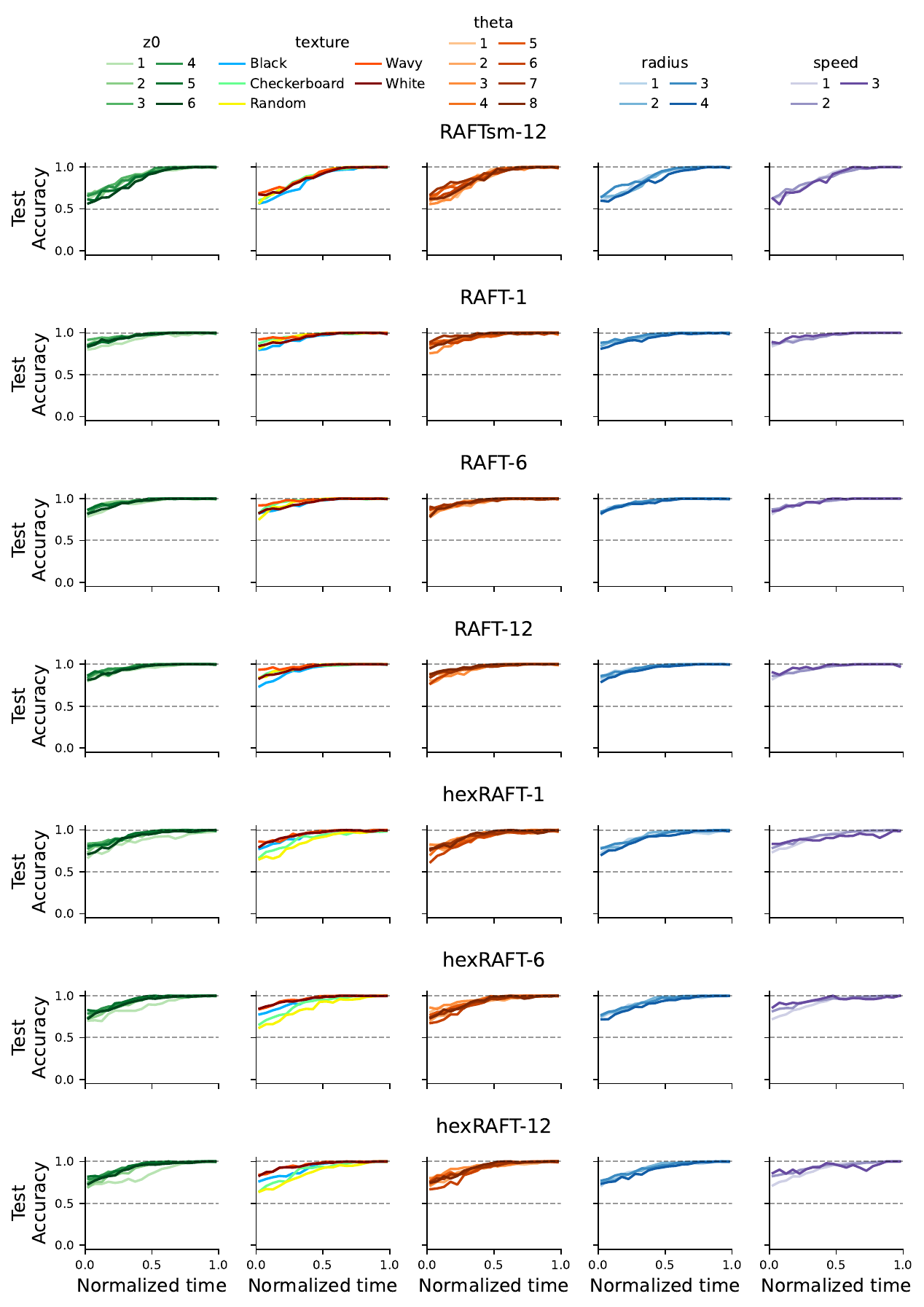}
    \caption{Decoder test accuracy over time for all RAFT networks, with normalized time (dividing by its total duration), conditioning on parameters of the simulation. Numbers in the legend correspond to ascending bins of the uniformly sampled parameters.}
    \label{fig_supp:decoder_test_accuracies_b}
\end{figure}

\clearpage
\subsubsection{Motion}
\begin{figure}[ht]
    \centering
    \includegraphics[width=\linewidth]{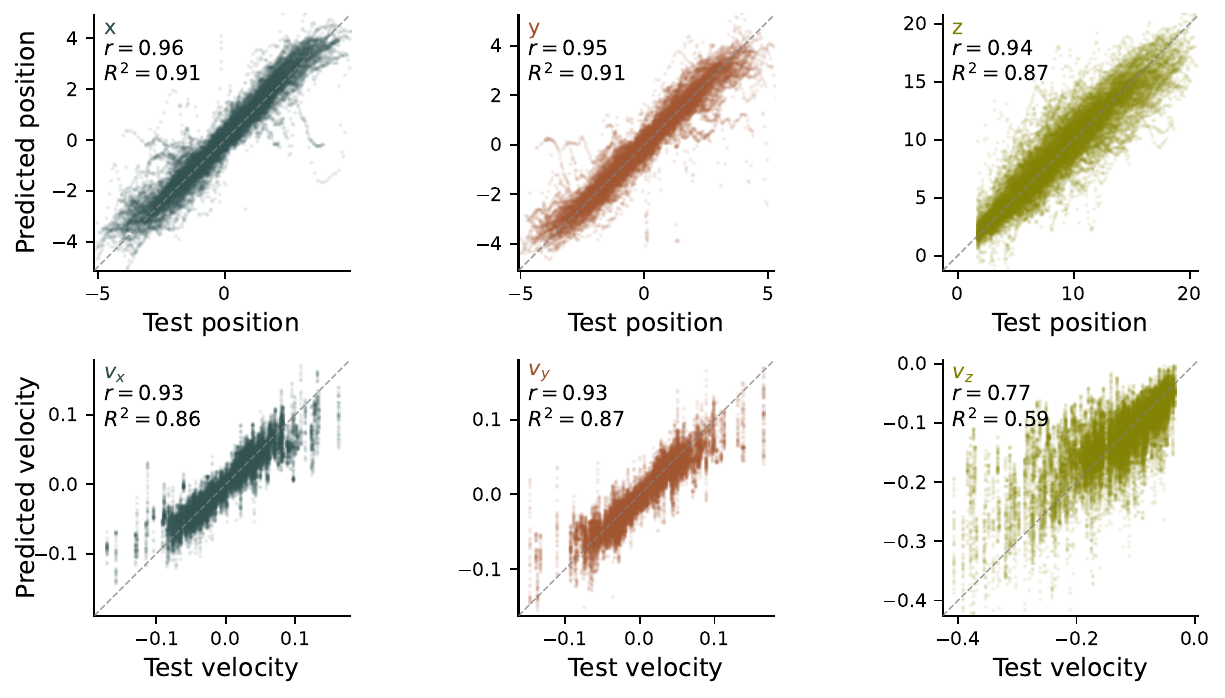}
    \caption{FlyNet decoder predictions versus ground truth for position (top) and velocity (bottom) on the test set. Columns correspond to the ($x$), ($y$), and ($z$) components. Pearson's ($r$) and ($R^2$) are reported.}
    \label{fig_supp:motion_decoders}
\end{figure}
\clearpage

\section{Additional results}

\subsection{Reverse-phi}

\begin{figure}[ht]
    \centering
    \includegraphics[width=\linewidth]{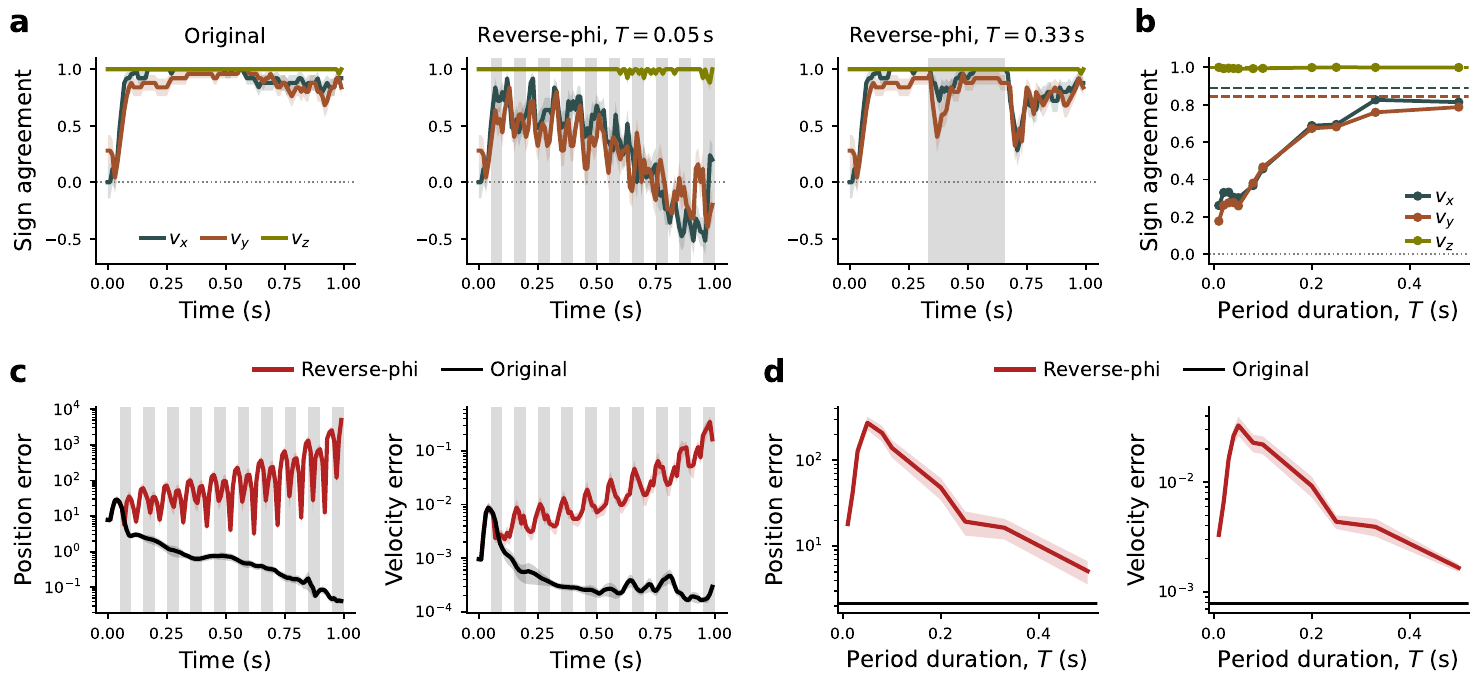}
    \caption{\textbf{FlyNet's representations of velocity consistently change sign during the reverse-phi.} \textbf{(a)} Sign agreement, i.e. the average product of true and predicted velocity sign across $N=50$ trajectories, given by $\frac{1}{N}\sum_N \text{sgn}({v_{\text{true}}}) \cdot \text{sgn}(\hat{v})$. Left: original trajectory. Middle: contrast inversion period of $T=0.05s$. Right: contrast inversion period of $T=0.33s$. Average and s.e.m. (shaded area) across 50 trajectories, gray shaded area represents the contrast flip. \textbf{(b)} Time-averaged sign agreement vs period duration $T$. Dashed lines represent the original trajectory values. \textbf{(c)} Time evolution of the position (left) and velocity (right) decoding error for reverse-phi with $T=0.05s$ inversion frames. \textbf{(d)} Time-averaged position (left) and velocity (right) decoding errors vs period duration $T$.}
    \label{fig_supp:reverse_phi_pos_vel_vs_T}
\end{figure}

\clearpage
\subsection{Parameter comparison across networks}

\begin{table}[ht]
    \centering
    \small
    \setlength{\tabcolsep}{4pt}

    \begin{tabular}{lcc}
        \toprule
        Network & Activations & Free parameters \\
        \midrule

        vCNN(2f)  & 276,480 & 414,064 \\
        vCNN(5f)  & 276,480 & 414,698 \\
        vCNN(10f) & 276,480 & 414,858 \\
        gCNN(2f)  & 276,480 & 414,064 \\

        RAFTsm-12 \citep{RAFT2020} & 432,640 & 1.0M \\
        RAFT-1 \citep{RAFT2020}    & 697,632 & 5.3M \\
        RAFT-6 \citep{RAFT2020}    & 697,632 & 5.3M \\
        RAFT-12 \citep{RAFT2020}   & 697,632 & 5.3M \\

        hexRAFT-1  & 264,192 & 5.3M \\
        hexRAFT-6  & 264,192 & 5.3M \\
        hexRAFT-12 & 264,192 & 5.3M \\

        FlyNet \citep{Lappalainen2024}
            & \textbf{45,669} & \textbf{734} \\

        CTRNN \citep{Lappalainen2024}
            & \textbf{45,669} & 11,593 \\

        sCTRNN
            & \textbf{45,669} & 1.51M \\

        \bottomrule
    \end{tabular}

    \caption{Parameter comparison across networks. In the case of sCTRNN, parameters comprise 1,513,231 per-edge weights, 65 resting potentials, and 65 time constants, yielding 1,513,361 free parameters.}
    \label{table:parameter_count}
\end{table}

\begin{figure}[ht]
    \centering
    \includegraphics[width=\textwidth]{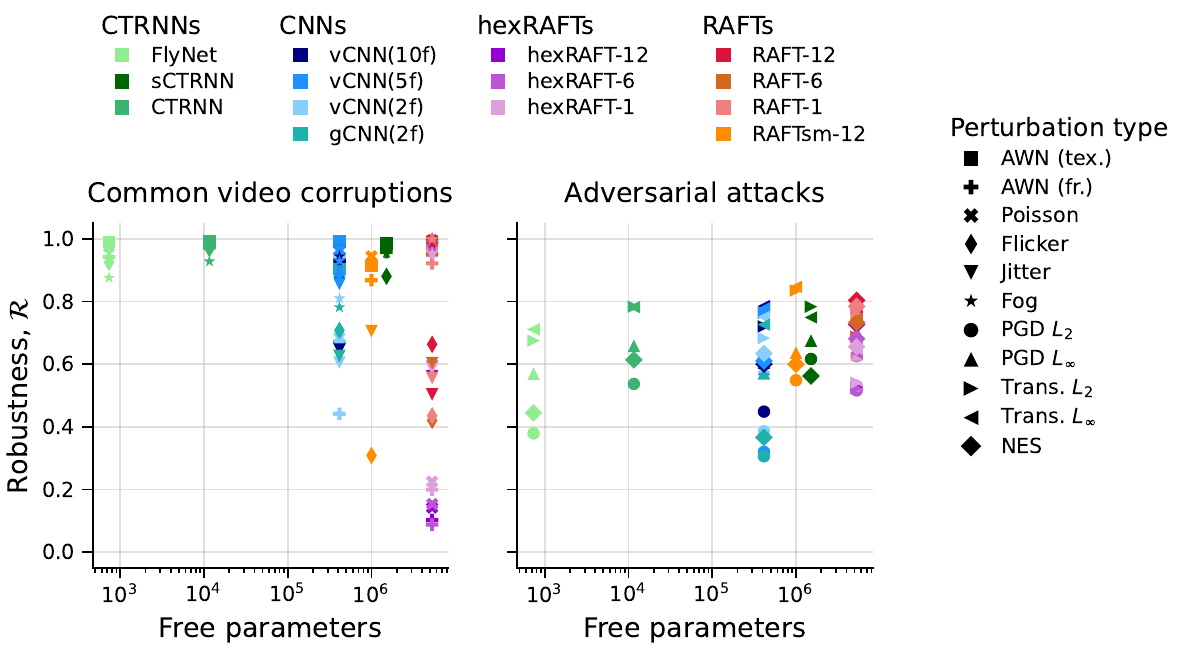}

    \caption{Robustness versus free parameters across models under common video corruptions (left) and adversarial attacks (right).}
    \label{fig:robustness_params}
\end{figure}

\newpage
\subsection{Optimization}
\subsubsection{Counterfactual restoration}
\begin{figure}[ht]
    \centering
    \includegraphics[width=\linewidth]{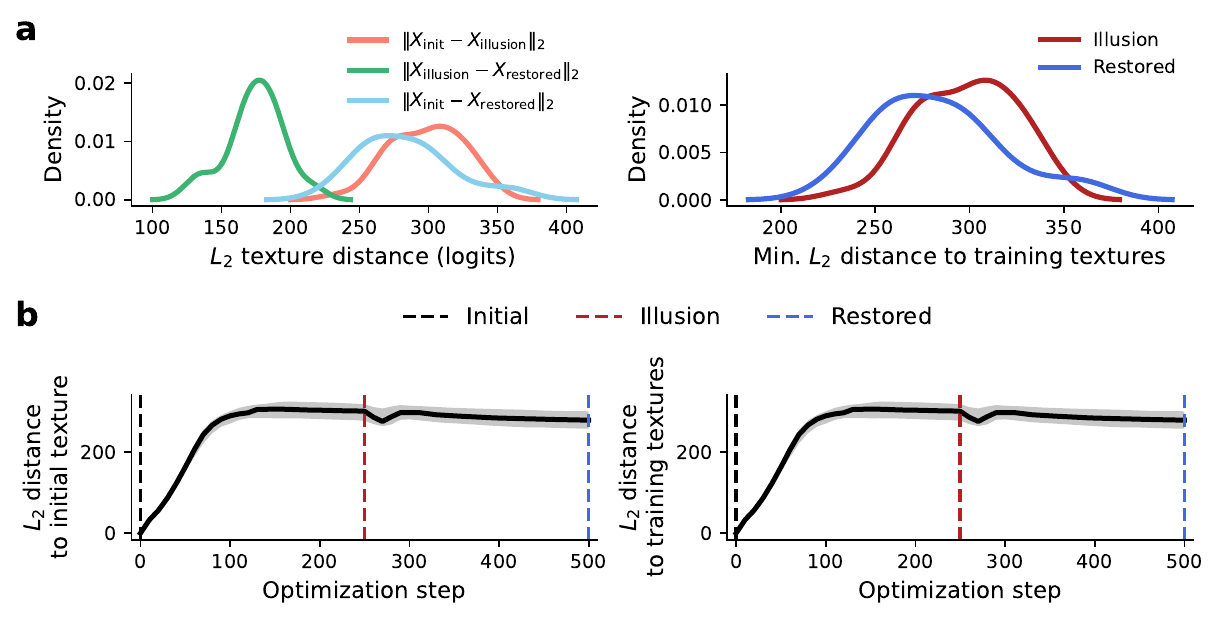}
    \caption{\textbf{$L_2$ distances of the textures during counterfactual restoration.} \textbf{(a)} $L_2$ distance distributions between the Initial, Illusion, and Restored textures (left), and the minimum distance between such textures and the training set textures (right). The base texture is sampled from the training set, so the distance is 0. \textbf{(b)} Same as \textbf{a} but median evolution over steps, with shaded IQR.}
    \label{fig_supp:distance_textures_counterfactual}
\end{figure}

\begin{figure}[ht]
    \centering
    \includegraphics[width=\linewidth]{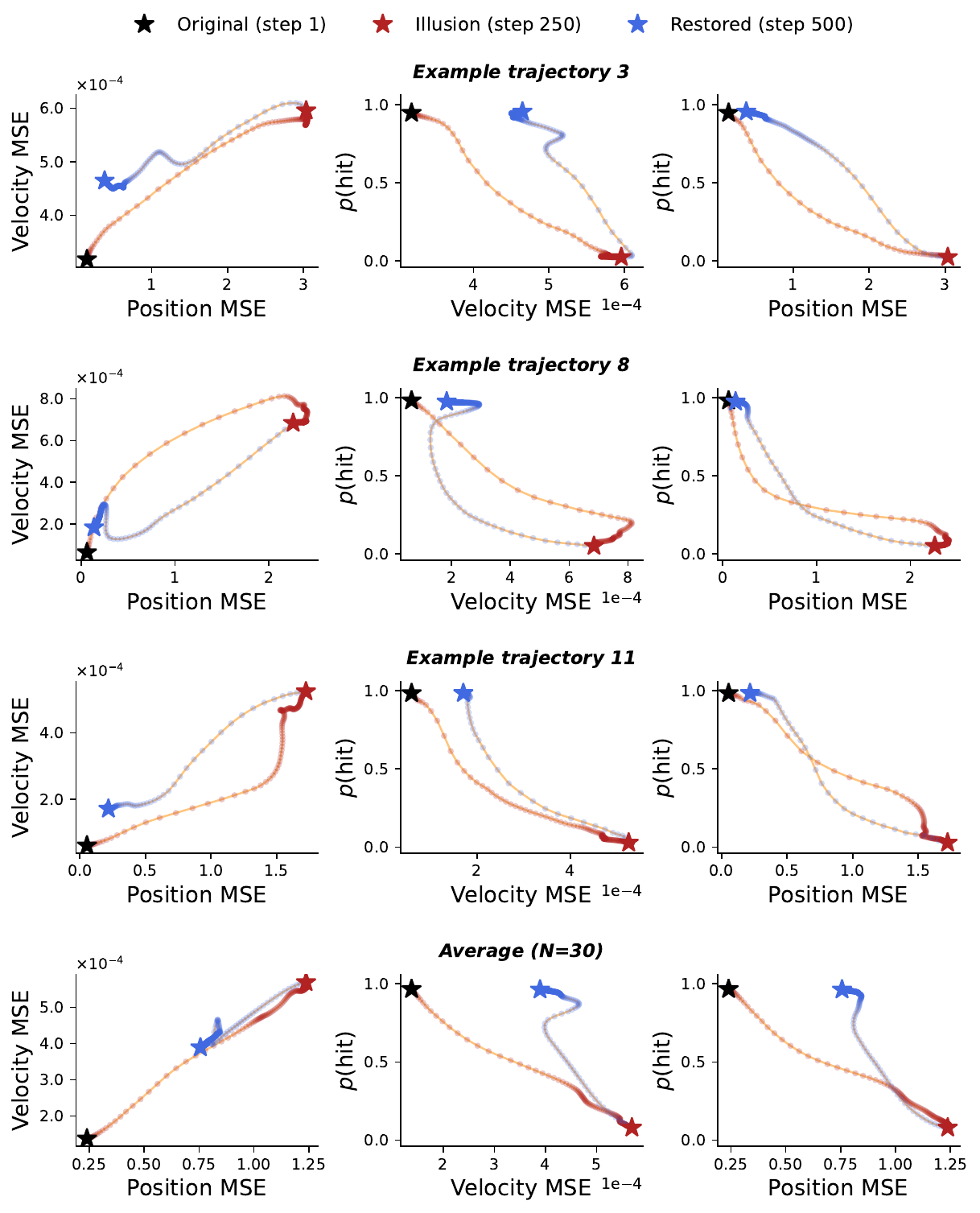}
    \caption{$p(\text{hit})$, position and velocity decoding errors over steps during counterfactual restoration. Top 3: Example trajectories. Bottom: Average across 30 trajectories.}
    \label{fig_supp:decoding_over_steps_counterfactual}
\end{figure}

\begin{figure}[ht]
    \centering
    \includegraphics[width=\linewidth]{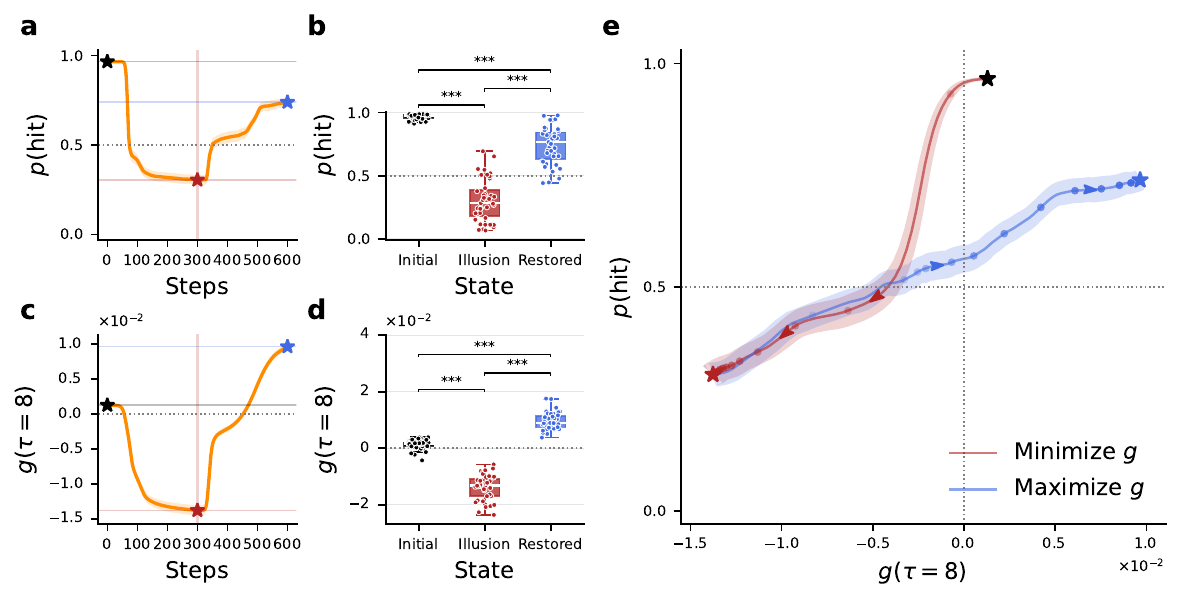}
    \caption{
        \textbf{Optimizing the input HR correlation alone produces the illusion and restores the hit response.} Textures were optimized to minimize $g(\tau=8)$ (Eq.~\ref{eq:hrc}) for 300 steps and then to maximize it for 300 steps, with a single Adam optimizer and an anchor penalty ($\lambda=0.1$; 30 trajectories). \textbf{(a)}~$p(\text{hit})$ across optimization steps. Stars mark initial (black), illusion (red), and restored (blue) textures. \textbf{(b)}~$p(\text{hit})$ at the three checkpoints. \textbf{(c)}~$g(\tau=8)$ across optimization steps. \textbf{(d)}~$g(\tau=8)$ at the three checkpoints. \textbf{(e)}~$p(\text{hit})$ against $g(\tau=8)$ along the optimization, during minimization (red) and maximization (blue); dot opacity and arrowheads indicate optimization steps. Lines show the mean; shaded area represents SEM; asterisks indicate significance from paired $t$-test (***: $p<0.001$).
        }
    \label{fig_supp:hr_correlator_failrestore}
\end{figure}

\begin{figure}[ht]
    \centering
    \includegraphics[width=\linewidth]{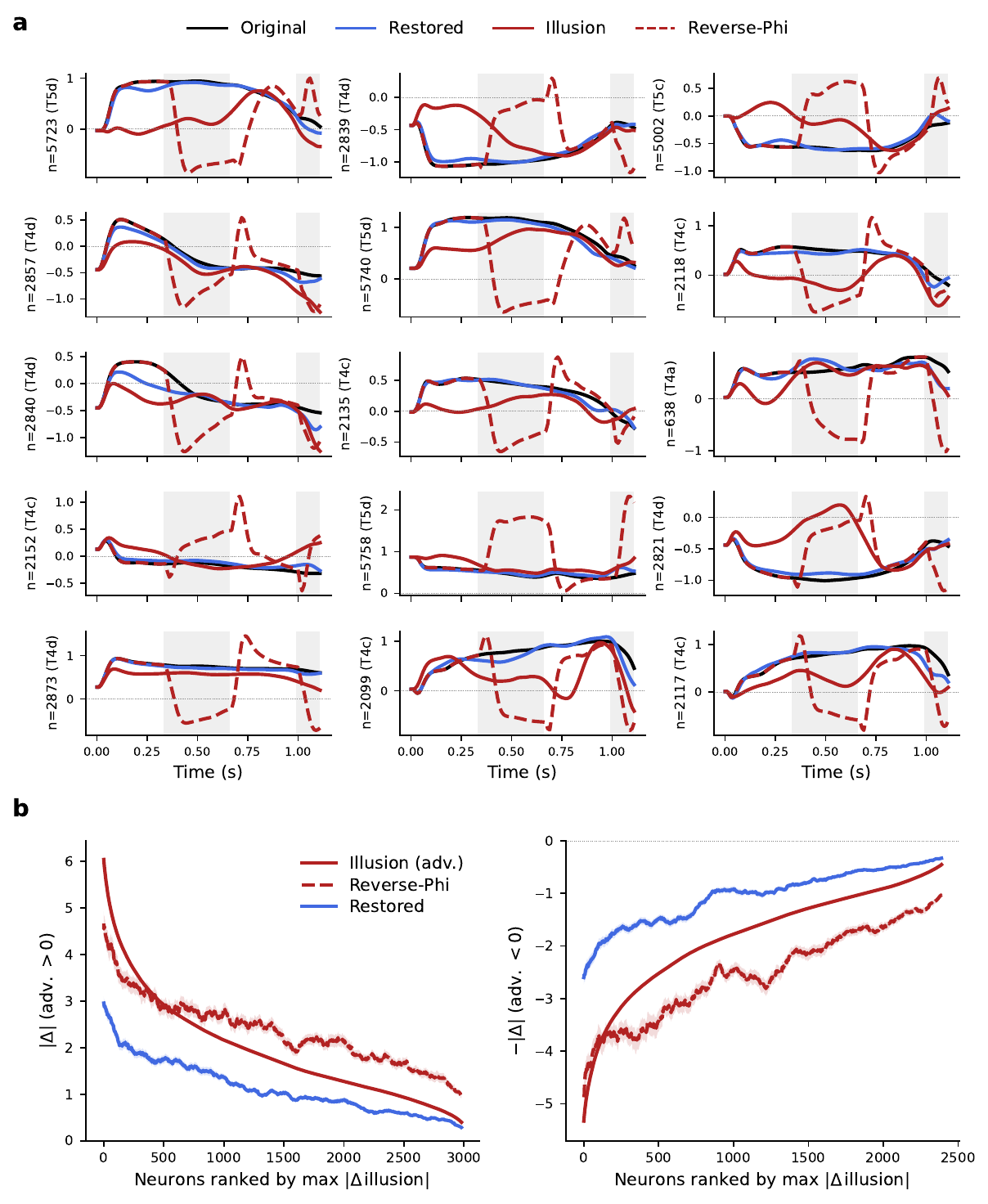}
    \caption{\textbf{(a)} Single neuron activity (each panel is a neuron) over time, for a trajectory, under all conditions: original texture, reverse-phi, illusion texture and restored texture (from counterfactual restoration). \textbf{(b)} Maximum single-neuron activity perturbation, defined as $\Delta=\text{Act. condition}-\text{Act. original}$. Taken across trajectories, computed in the window of contrast change for reverse-phi ---for reverse-phi, illusion and restored textures (from counterfactual restoration)---. Neurons sorted according to the maximum perturbation given by illusory texture. Left: predicted positive by the illusory texture. Right: predicted negative by the illusory texture. Mean rolling window of 200 neurons for smoothing.}
    \label{fig_supp:tracesndisruption}
\end{figure}

\clearpage
\subsection{Representational Similarity Analysis (RSA)}\label{app:rsa}
We use representational similarity analysis \citep{kriegeskorte2008rsa} to quantify how each condition's T4/T5 population response relates to reverse-phi. We represent every condition by the window-mean activity of its $5768$ T4/T5 units over the full $330$~ms first-contrast-flip window, and compare these population vectors, $\textbf{r}_i$, taking reverse-phi as the reference. We use three metrics: the Pearson correlation across units, the Euclidean distance (their $\ell_2$ distance normalized by $\sqrt{n_\mathrm{units}}$), and a signed projection.

\begin{figure}[ht]
    \centering
    \includegraphics[width=\linewidth]{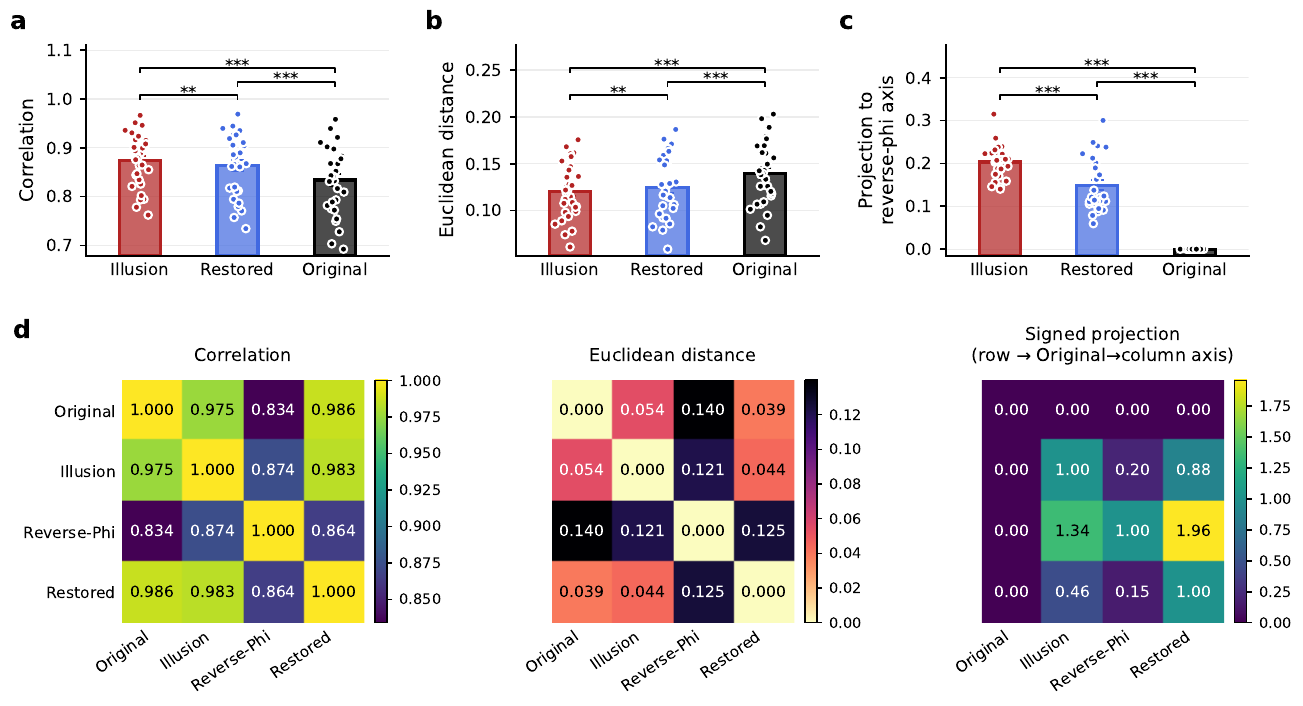}
    \caption{\textbf{Representational similarity to reverse-phi.}
    \textbf{(a--c)} Closeness of each condition's T4/T5 population response to reverse-phi, under three metrics: Pearson correlation across units \textbf{(a)}, Euclidean distance \textbf{(b)}, and the signed projection onto the original$\rightarrow$reverse-phi axis \textbf{(c)}, on which the original sits at 0 and reverse-phi at 1. Bars are means over the 30 trajectories, error bars s.e.m., dots individual trajectories; brackets give paired $t$-tests (\textsuperscript{**}$p<0.01$, \textsuperscript{***}$p<0.001$). The adversarial texture is closer to reverse-phi than the restored one on every metric.
    \textbf{(d)} The full condition-by-condition matrix for each metric; panels \textbf{a--c} are its reverse-phi column. Each matrix is computed within a trajectory and averaged over the 30 trajectories. In \textit{signed projection} entry $(i,j)$ gives how far condition $i$ lies along the original$\rightarrow j$ axis, so the original row and column are 0 by construction; reverse-phi overshoots the adversarial texture along the original$\rightarrow$adversarial axis (1.34), placing the adversarial
    response between the original and reverse-phi rather than beyond it.}
    \label{fig_supp:rsa}
\end{figure}

Relative to reverse-phi, the adversarial representation showed higher correlation (0.874) than the restored (0.864) and original (0.834) representations, and lower Euclidean distance (0.121 vs.\ 0.125 and 0.140, respectively; Supp. Fig.~\ref{fig_supp:rsa}a,b). A signed projection onto the original--reverse-phi axis further confirmed this shift (0.203 for adversarial vs.\ 0.149 for restored, with original defined as 0; Supp. Fig.~\ref{fig_supp:rsa}c).

The signed projection computes
\begin{equation*}
S_{ij} \;=\; \frac{\lVert \mathbf{d}_i \rVert}{\lVert \mathbf{d}_j \rVert}\,\cos\theta_{ij},
\end{equation*}
the scalar projection of $\textbf{d}_i$ onto $\textbf{d}_j$, where $\mathbf{d}_i = \mathbf{r}_i - \mathbf{r}_\mathrm{orig}$ is the displacement of condition $i$ from the original. The original sits at $0$ and condition $j$ at $1$. $\theta_{ij}$ is the angle between the two representation displacements.  The headline comparison (Supp. Fig.~\ref{fig_supp:rsa}c) takes $j=\,$reverse-phi.

Both adversarial and restored align with reverse-phi above chance ($\theta\approx58^\circ$). Their shifts differ in magnitude: the adversarial shift is the longer of the two, so it projects farther onto the original$\rightarrow$reverse-phi axis ($0.20$ vs.\ $0.15$) while reverse-phi overshoots both ($1.34$, $1.96$), placing the adversarial response between the original and reverse-phi and closer to reverse-phi than the restored one. This is consistent with single-neuron traces (Supp. Fig.~\ref{fig_supp:tracesndisruption}a), where the illusion response departs from the original in the same direction as reverse-phi but at reduced amplitude, while the restored response tracks the original.

All three metrics showed significant differences between the adversarial representation and both the restored and the original one (paired $t$-tests across the 30 trajectories, all $p<0.005$). The condition-by-condition matrices for the three metrics are shown in Supp. Fig.~\ref{fig_supp:rsa}d--f.
\clearpage
\subsection{HR correlator}
\begin{figure}[ht]
    \centering
    \includegraphics[width=\linewidth]{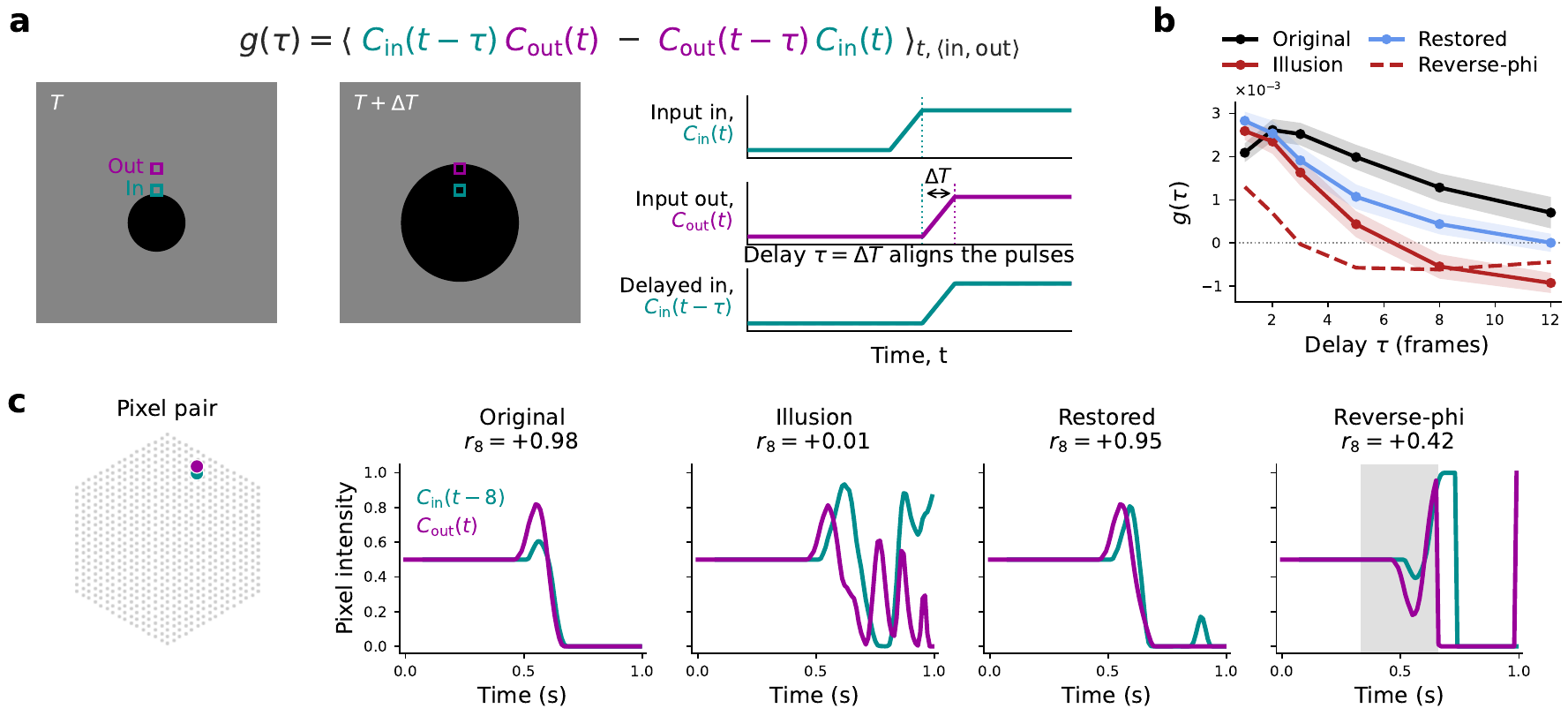}
    \caption{\textbf{The Hassenstein--Reichardt looming correlation.} \textbf{(a)} Cartoon, two consecutive frames of a looming object (black ball) approaching the camera. Two neighboring pixels $(in, out)$ are highlighted, with the activity of the highlighted neighboring pixels, and a delayed example. \textbf{(b)} Dependence of $g(\tau)$ on $\tau$ for FlyNet, for the 4 conditions (all averaged at the same window as reverse-phi). Average across 30 trajectories, shade indicates standard error of the mean. \textbf{(c)} Real example, pixel pair highlighted. Traces of the pixels ($in$ delayed) for the 4 conditions: Original, Adversarial illusion, Restored, Reverse-phi. Pearson correlation of the traces is reported on top.}
    \label{fig_supp:hrc_cartoon}
\end{figure}

\begin{figure}[ht]
    \centering
    \includegraphics[width=\linewidth]{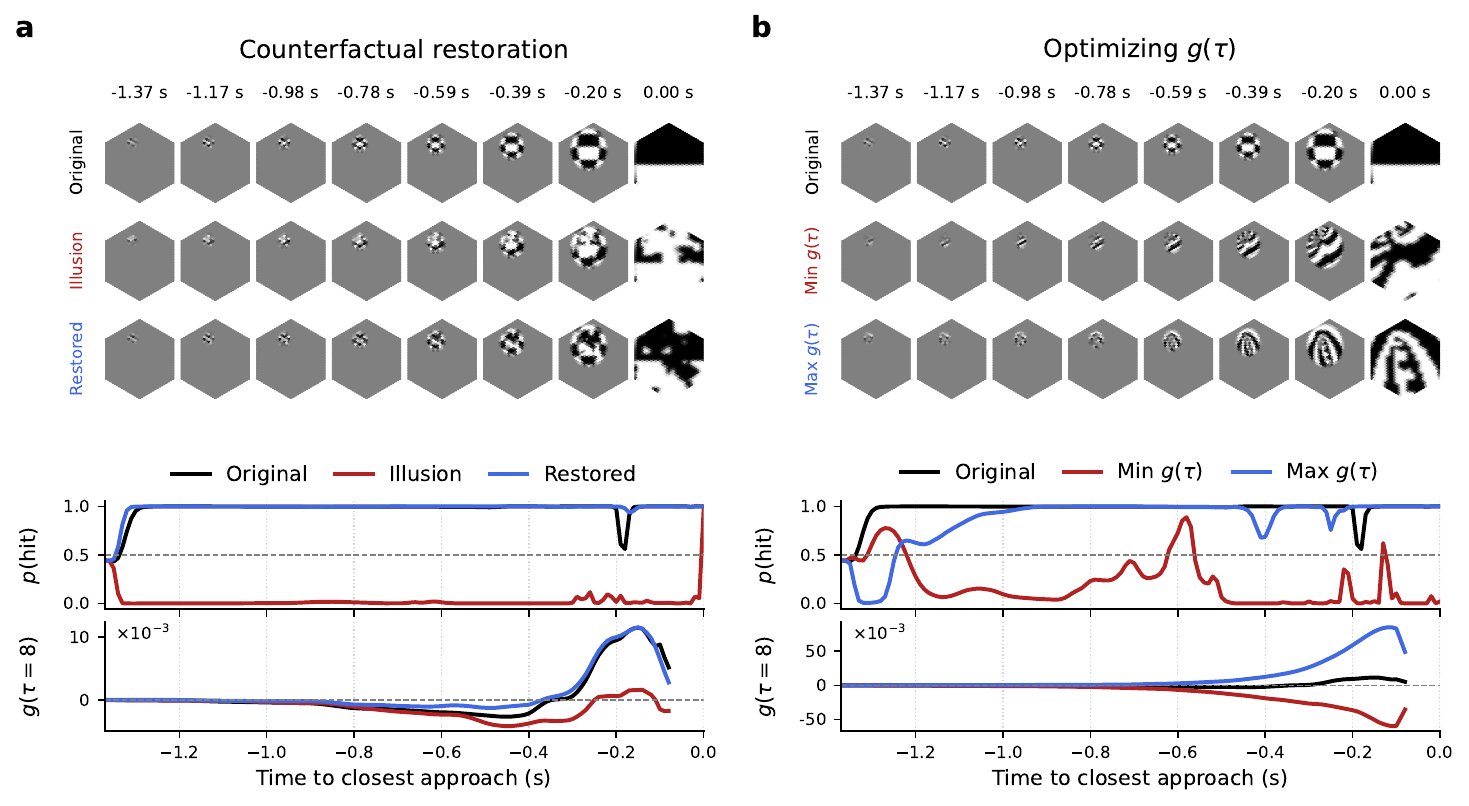}
    \caption{\textbf{Comparison between counterfactual restoration and $g(\tau)$ optimization on FlyNet.} Top hex-panels: evolution over time of the input for the 3 conditions. Middle panel: $p(\text{hit})$ over time. Bottom panel: $g(\tau)$ over time. \textbf{(a)} Counterfactual restoration: original (black), adversarial illusion (red) and restored (blue). \textbf{(b)} Optimizing $g(\tau)$: original (black), minimizing (red) and maximizing (blue) $g(\tau)$. }
    \label{fig_supp:comparison_phit_gtau}
\end{figure}
\clearpage
\subsection{Attacks}
\subsubsection{Adversarial attacks}\label{app:adv_attacks}

\paragraph{Methods} We perform untargeted adversarial attacks over the object texture by optimizing a free logit variable, ensuring valid pixel intensities through a sigmoid transformation. All adversarial evaluations are performed on 100 time-steps hit trajectories with Black initial texture where baseline accuracy exceeds 90\%, ensuring that failures arise from perturbations rather than pre-existing model errors. For PGD \citep{Madry2018}, we iteratively update the texture to maximize misclassification under $L_\infty$ and $L_2$ constraints via projection, using 100 steps with step size $\alpha=\varepsilon/10$. During optimization, we employ an adaptive step-size scheme inspired by \citep{croce2020autoattack}: at fixed checkpoints of 20 steps, if the loss fails to improve, the step size is halved and the iterate is reset to the best solution found so far; otherwise, optimization proceeds with the current step size. We sweep over $\varepsilon_{L_\infty} = [1, 6, 8, 10, 15]$ and $\varepsilon_{L_2} = [100, 250, 500, 1000, 2000]$. Recall we optimize the texture in logit space. Each attack uses 2 random restarts from noise within the constraint set, and we run 15 independent attacks per condition. We report the highest label-flip rate across restarts. We then use the resulting PGD adversarial textures to assess transfer by applying them directly to the other models. For CW attacks \citep{carliniwagner2017}, we optimize the same variable using Adam to minimize a distance term plus a misclassification objective, performing a grid search over $c = [1, 10, 10^2, 10^3, 10^4, 10^5]$ with 150 steps and $\text{lr}=10^{-1}$, evaluating both $L_\infty$ and $L_2$ distances. Finally, for the gradient-free NES method \citep{wierstra2008nes}, we estimate the update direction using symmetric Gaussian perturbations, sweeping over $\sigma = [1, 3, 6, 9, 12]$, using 100 optimization steps with 15 gradient-estimation samples per step and a normalized update, and running 15 attacks per condition.

\begin{figure}[ht]
    \centering
    \includegraphics[width=\linewidth]{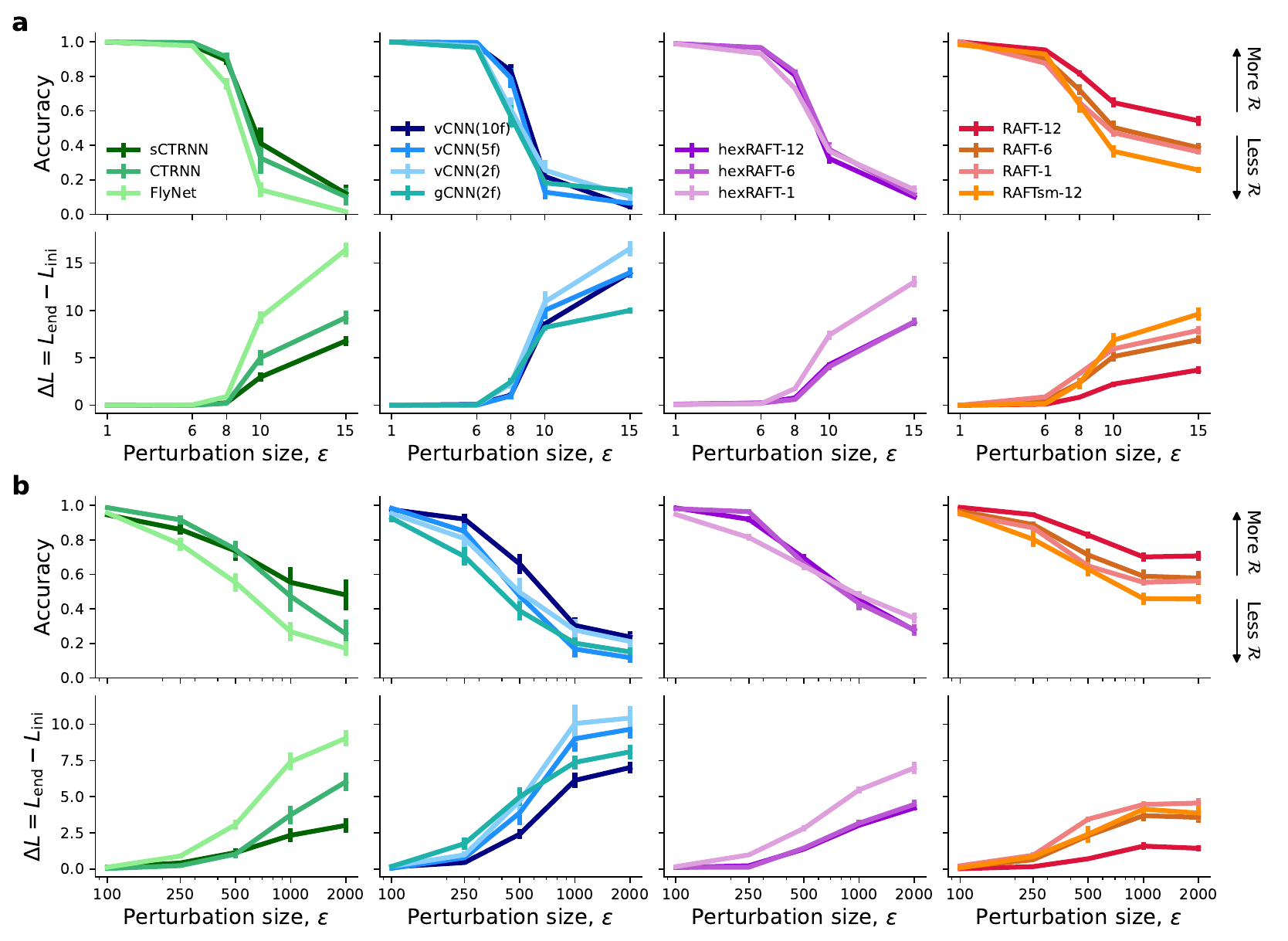}
    \caption{\textbf{Robustness of models under projected gradient descent (PGD) attacks.} (a) PGD-$L_\infty$ and (b) PGD-$L_2$ attacks across model architectures. Classification accuracy and change in loss ($L = L_{\text{end}} - L_{\text{ini}}$) are shown as functions of perturbation size $\varepsilon$. Error bars indicate variability across samples.}
    \label{fig_supp:adversarial_curves}
\end{figure}

\begin{figure}[ht]
    \centering
    \includegraphics[width=\linewidth]{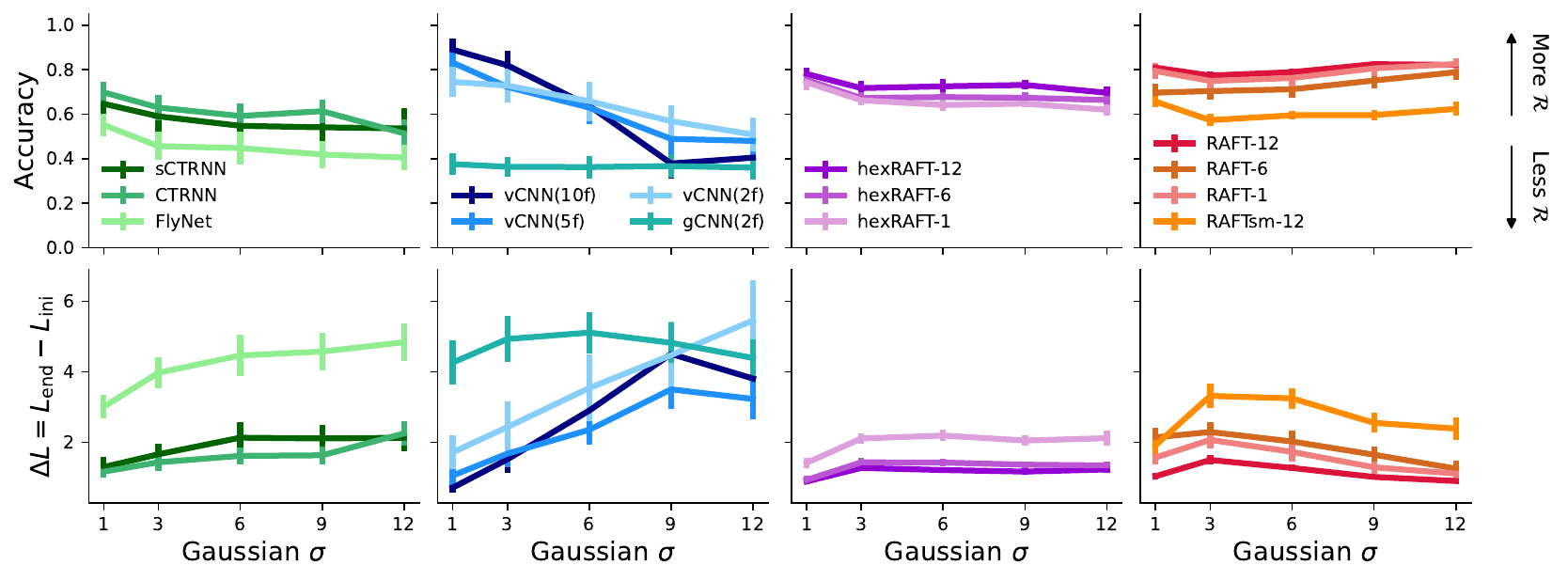}
    \caption{\textbf{Robustness of models under natural evolution strategies (NES) attacks.} Classification accuracy and change in loss ($L = L_{end} - L_{ini}$) are shown as functions of the Gaussian perturbation standard deviation $\sigma$ across model architectures. Error bars indicate variability across samples.}
    \label{fig_supp:adversarial_NES}
\end{figure}

\begin{figure}
    \centering
    \includegraphics[width=\linewidth]{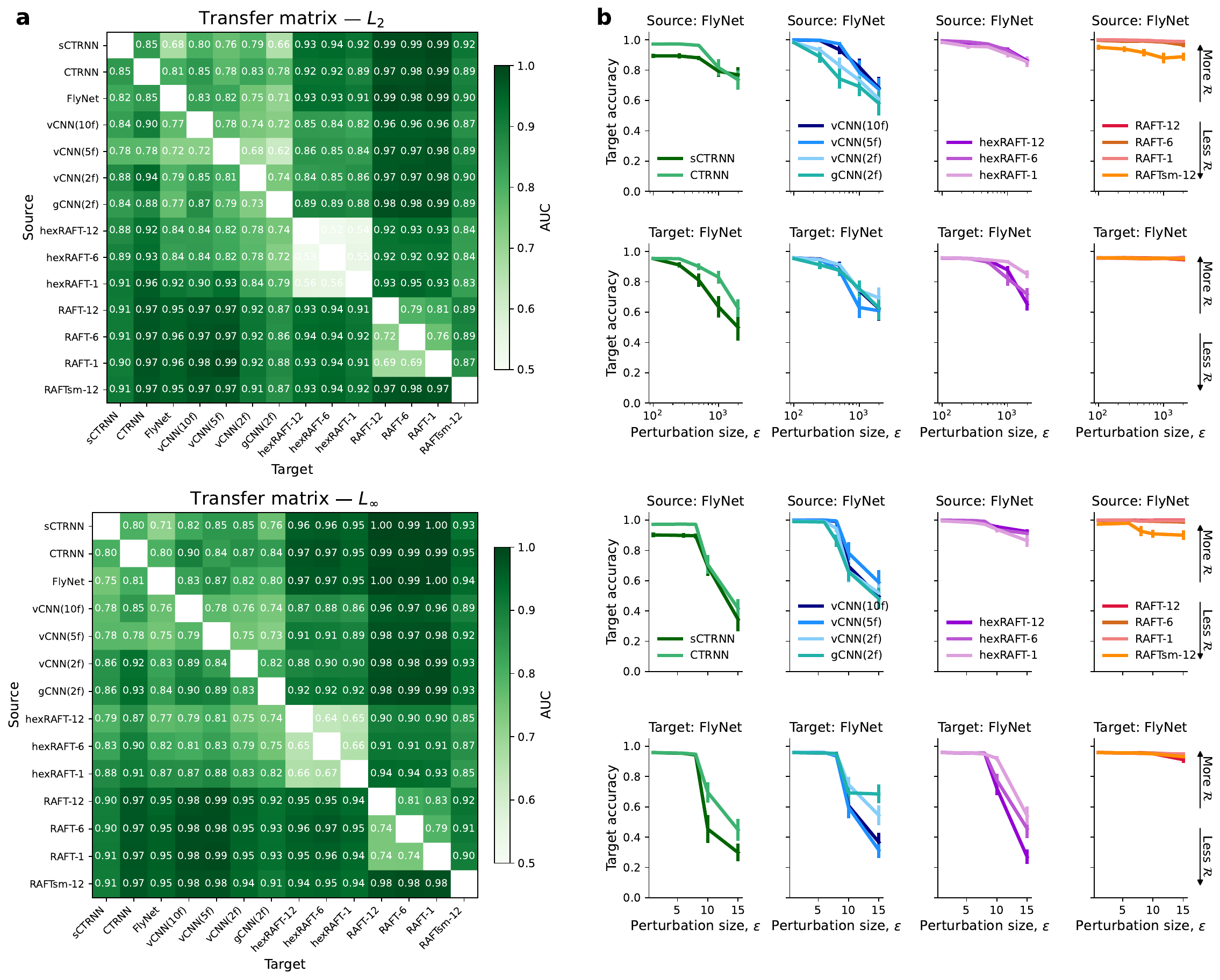}
    \caption{\textbf{(a)} Transfer attack AUC matrices for PGD illusions. Rows: Source of illusion. Columns: Target of the illusion. Top: $L_2$ norm. Bottom: $L_\infty$ norm. \textbf{(b)} Transfer attack curves for flynet as source (left) and target (right). Top: $L_2$ norm. Bottom: $L_\infty$ norm.}
    \label{fig_supp:transfer_matrices}
\end{figure}

\begin{figure}
    \centering
    \includegraphics[width=\linewidth]{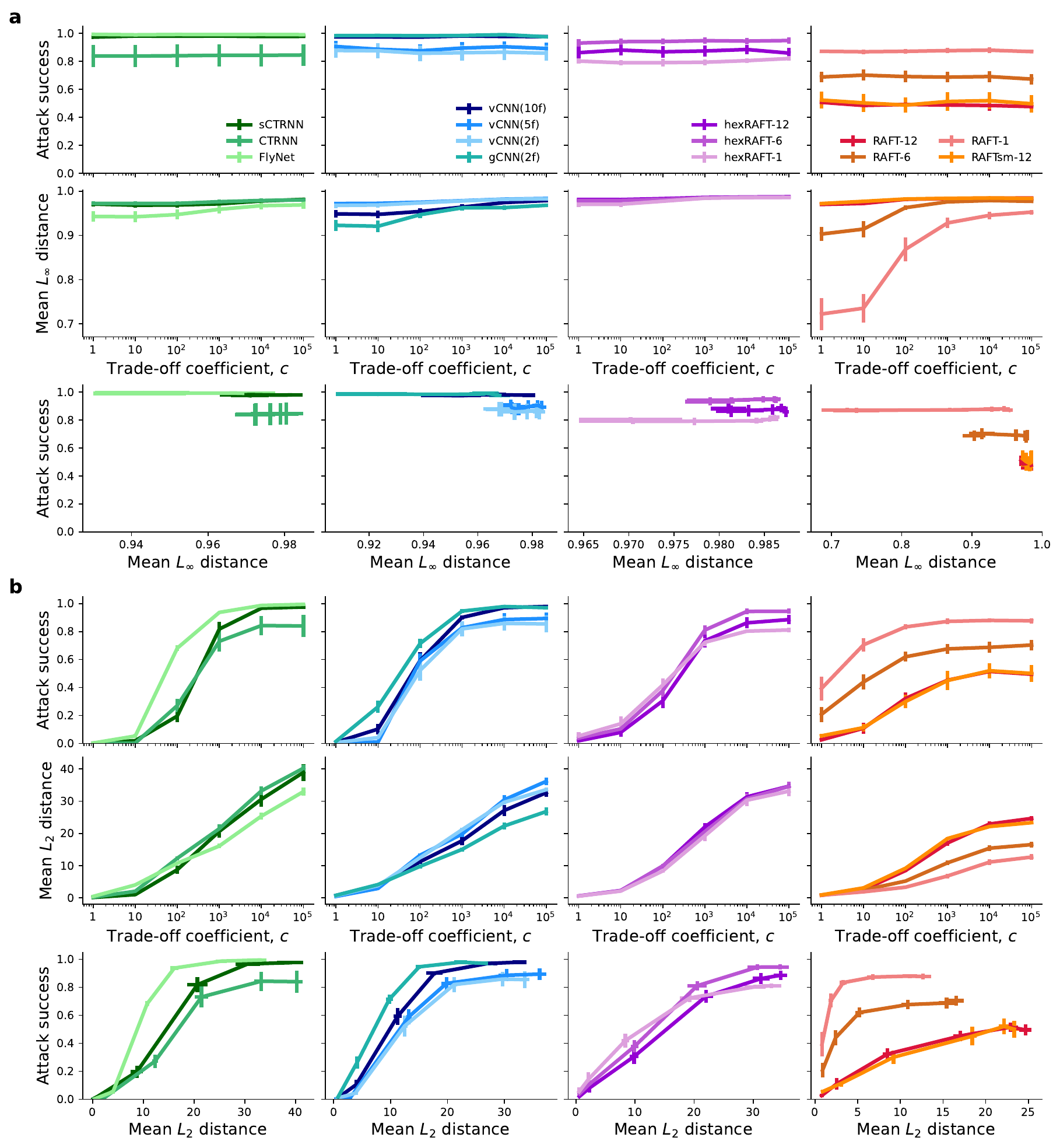}
    \caption{\textbf{Robustness of models under Carlini--Wagner (CW) attacks.} (a) CW-$L_\infty$ and (b) CW-$L_2$ attacks across model architectures. Attack success rates and mean perturbation distances are shown as functions of the attack parameter (c), together with attack success as a function of perturbation distance. Error bars indicate variability across samples.}    
    \label{fig_supp:cw_curves}
\end{figure}

\begin{figure}[ht]
    \centering
    \includegraphics[width=\linewidth]{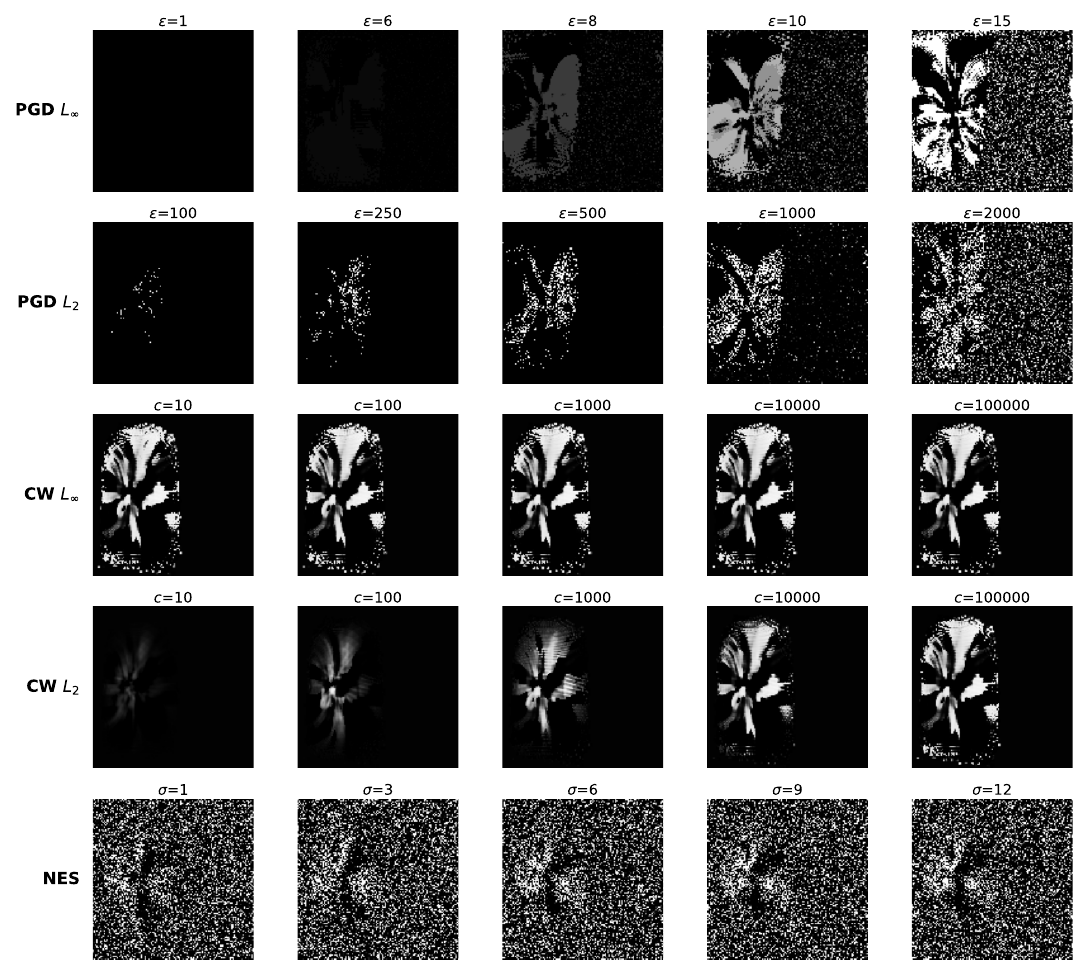}
    \caption{\textbf{Examples of adversarial textures found for FlyNet.}
    Adversarial textures optimized for the synthesized FlyNet model under PGD, CW, and NES attacks. Columns show increasing perturbation strength, either $\varepsilon$ for PGD, $c$ for CW, or $\sigma$ for NES, illustrating the distinct texture structures induced by each attack objective and constraint.}
    \label{fig_supp:flynet_adversarial_examples}
\end{figure}

\clearpage
\subsubsection{Common video corruptions}\label{app:video_perturbations}
\textit{Additive White Noise (AWN) on Texture}: It simulates sensor noise or surface irregularities on the looming object. All frames are given by adding zero-mean independent white noise with variance $\sigma^2$ to the texture. 

\textit{AWN on Frames}: We test robustness to global unstructured perturbations across the entire visual field, simulating temporal sensor grain and atmospheric degradation. At each frame, we add zero-mean independent white noise with variance $\sigma^2$ to every input pixel. For FlyNet, CNNs, and hexRAFT we add it to every pixel of the hexagonal input of $N_\text{FlyNet}=721$ columns $V_t$, for RAFT to every pixel of the full image $F_t$. Noise power is matched across architectures by scaling $\sigma(\text{arch.})=\sigma \sqrt{N_\text{FlyNet} / N_{\text{arch.}}}$. 

\textit{Shot (Poisson) noise}: Similar to the previous perturbation but with normalized Poisson noise instead of Gaussian, modulated by the photon count $c$. Noise power is matched across architectures by scaling the photon count as $c(\text{arch.}) = c \cdot N_{\text{arch.}}/N_{\text{FlyNet}}$. 

\textit{Frame Flickering}: We apply random, frame-wise gain fluctuations ($G_t = 1+ \xi_t$), modulated by the flickering strength $\varepsilon$ as $\xi_t \sim \mathcal{U}(-\varepsilon, \varepsilon)$. 

\textit{Motion jitter}: We apply random, frame-wise shifts of the whole image in both vertical and horizontal dimensions. We change the maximum amount of shifted pixels by the jitter $j$ magnitude. 

\textit{Contrast Sensitivity (Fog)}: We reduce stimulus contrast against the background ($\hat{\mathbf{F}} = (\mathbf{F}-b)\alpha + b$). $\alpha$ modulates the amount of contrast scale difference between the object and the background, $\alpha=1$ is the original video and $\alpha=0$ are just gray frames.

Results for AWN, shot noise, flickering, and motion jitter are averaged over 10 trajectories with 10 noise realizations per trajectory, while fog results are averaged over 20 trajectories.

\begin{figure}[ht]
    \centering
    \includegraphics[width=\linewidth]{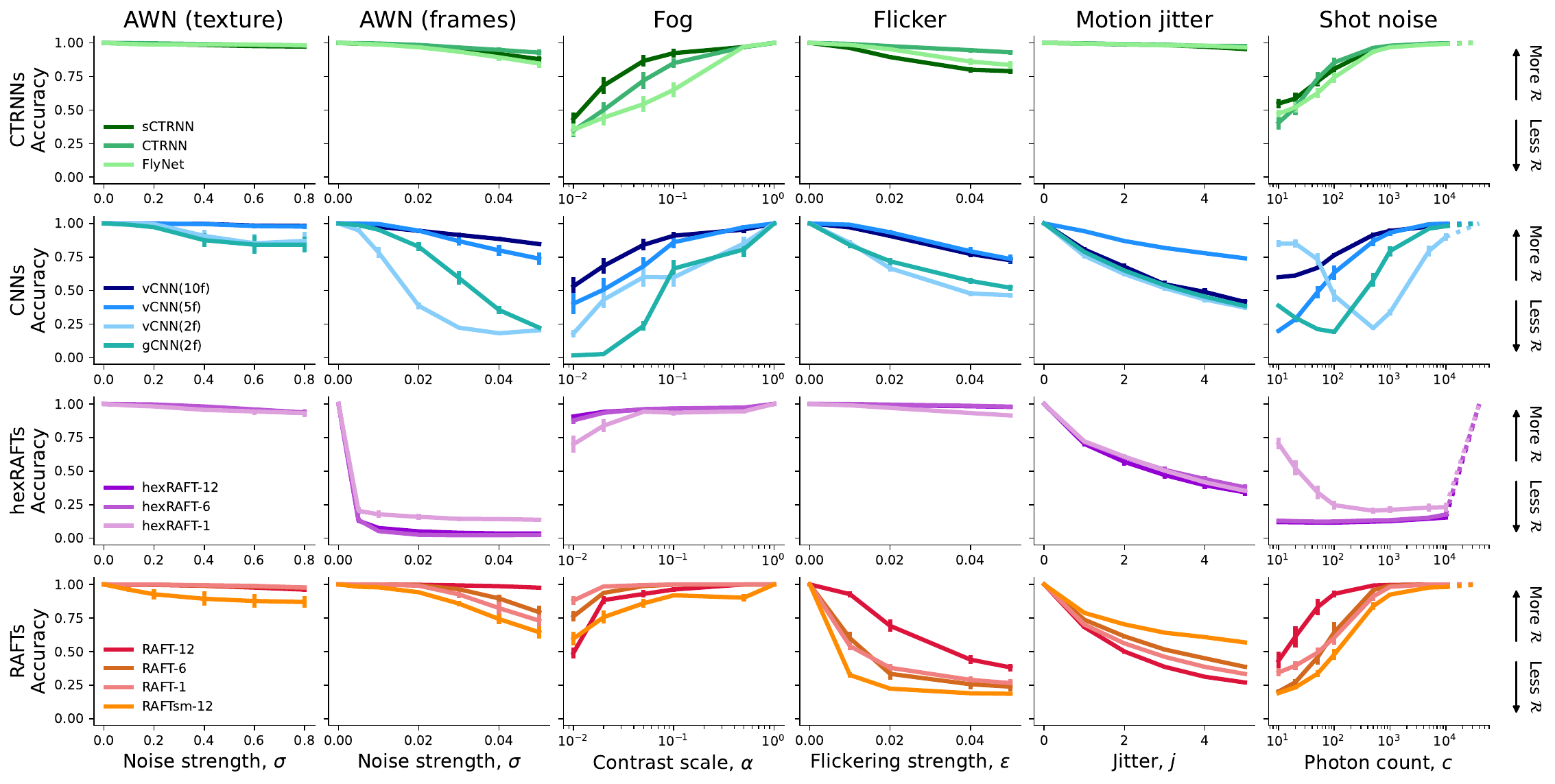}
    \caption{\textbf{Accuracy of networks under common video perturbations, for each network family.} Order (columns): Additive White Noise (AWN) on the texture, AWN on the frames, Fog/Reduced contrast, Flicker, Motion Jitter and Shot (Poisson) noise. The first five consist on the average across 10 trajectories and 10 noise samples for each trajectory. The last consists on the average across 20 trajectories. Each row corresponds to a network family.}
    \label{fig_supp:environmental_stressors_curves}
\end{figure}

\begin{figure}[ht]
    \centering
    \includegraphics[width=1\linewidth]{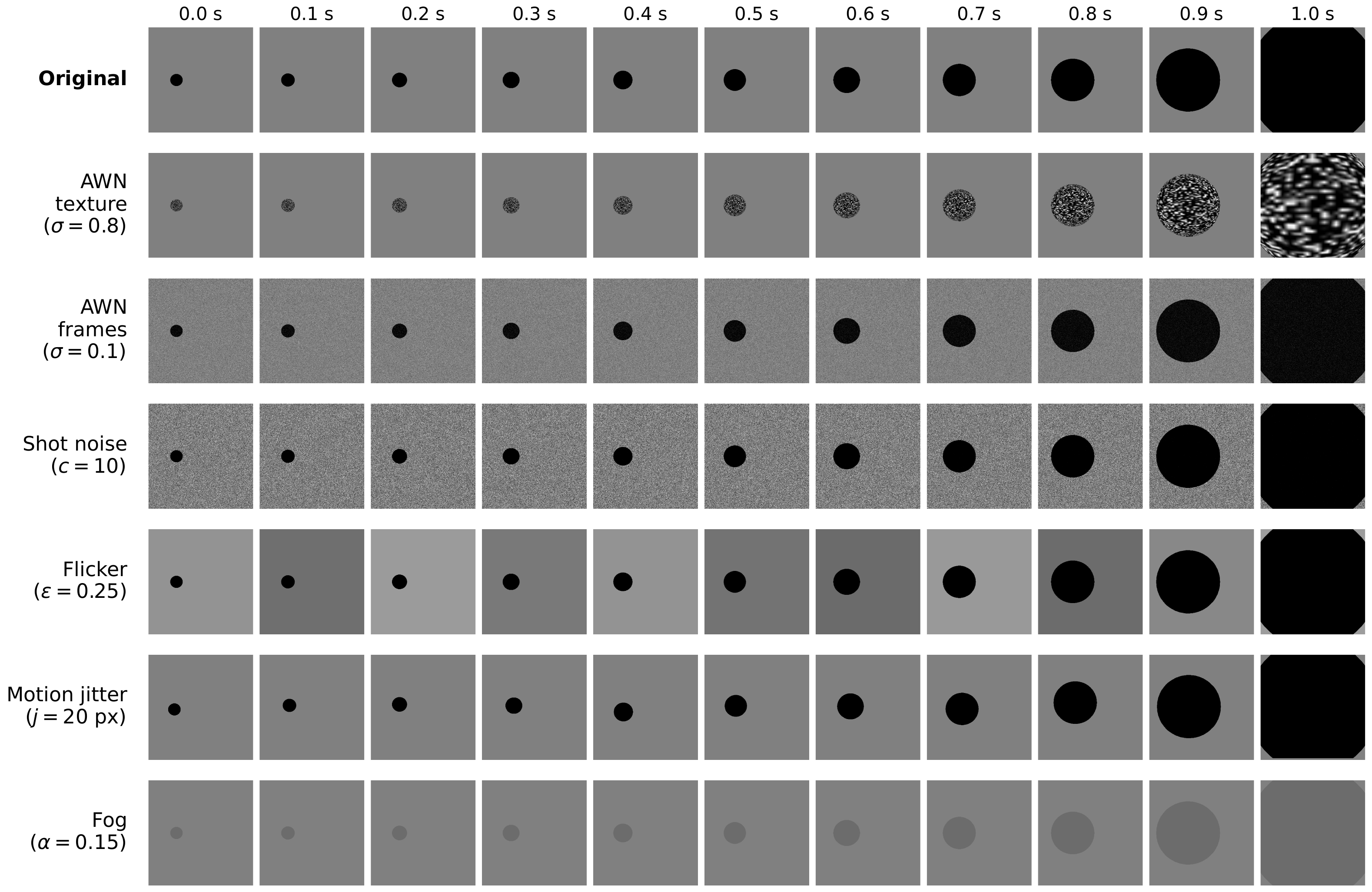}
    \caption{Examples of frames over time of each different video corruption.}
    \label{fig_supp:environmental_stressors_examples}
\end{figure}

\clearpage

\begin{landscape}
\subsubsection{Summary table}
\begin{table*}[h]
\newcommand{\tblbest}[1]{\underline{\textbf{\boldmath #1}}}
\newcommand{\tblsecond}[1]{\textbf{\boldmath #1}}
\newcommand{\tblthird}[1]{\textit{#1}}
\centering
\setlength{\tabcolsep}{3pt}%
\renewcommand{\arraystretch}{1.12}%
\resizebox{\linewidth}{!}{%
\begin{tabular}{lll*{14}{c}}
\toprule
 & & & {FlyNet} & {sCTRNN} & {CTRNN} & {vCNN(10f)} & {vCNN(5f)} & {vCNN(2f)} & {gCNN(2f)} & {hexRAFT-12} & {hexRAFT-6} & {hexRAFT-1} & {RAFT-12} & {RAFT-6} & {RAFT-1} & {RAFTsm-12} \\
\midrule
\multicolumn{3}{l}{$\mathcal{E}$ (GFLOPs)} & 0.186 & \textbf{0.183} & 0.783 & 71.7 & 75.6 & 78.8 & 78.8 & 10,290 & 6,495 & 3,332 & 27,402 & 17,380 & 9,029 & 6,138 \\
\midrule
\multirow{12}{*}{\rotatebox[origin=c]{90}{\textbf{Adversarial attacks}}} & \multirow{2}{*}{PGD $L_2$} & $\mathcal{R}$ & 0.379 $\pm$ 0.040 & 0.616 $\pm$ 0.068 & 0.537 $\pm$ 0.058 & 0.448 $\pm$ 0.032 & 0.320 $\pm$ 0.034 & 0.385 $\pm$ 0.052 & 0.306 $\pm$ 0.022 & 0.524 $\pm$ 0.019 & 0.516 $\pm$ 0.026 & 0.531 $\pm$ 0.019 & \tblbest{0.766 $\pm$ 0.021} & \tblsecond{0.657 $\pm$ 0.033} & \tblthird{0.625 $\pm$ 0.014} & 0.548 $\pm$ 0.031 \\
 &  & LERS & 0.046 $\pm$ 0.005 & \tblbest{0.075 $\pm$ 0.008} & \tblsecond{0.060 $\pm$ 0.007} & 0.041 $\pm$ 0.003 & 0.029 $\pm$ 0.003 & 0.035 $\pm$ 0.005 & 0.028 $\pm$ 0.002 & 0.040 $\pm$ 0.001 & 0.040 $\pm$ 0.002 & 0.042 $\pm$ 0.002 & \tblthird{0.057 $\pm$ 0.002} & 0.050 $\pm$ 0.002 & 0.048 $\pm$ 0.001 & 0.043 $\pm$ 0.002 \\
 & \multirow{2}{*}{PGD $L_\infty$} & $\mathcal{R}$ & 0.569 $\pm$ 0.014 & \tblthird{0.675 $\pm$ 0.029} & 0.658 $\pm$ 0.027 & 0.605 $\pm$ 0.009 & 0.584 $\pm$ 0.015 & 0.592 $\pm$ 0.027 & 0.571 $\pm$ 0.012 & 0.632 $\pm$ 0.009 & 0.651 $\pm$ 0.015 & 0.631 $\pm$ 0.013 & \tblbest{0.793 $\pm$ 0.015} & \tblsecond{0.703 $\pm$ 0.018} & 0.674 $\pm$ 0.012 & 0.636 $\pm$ 0.017 \\
 &  & LERS & \tblthird{0.069 $\pm$ 0.002} & \tblbest{0.082 $\pm$ 0.003} & \tblsecond{0.074 $\pm$ 0.003} & 0.056 $\pm$ 0.001 & 0.054 $\pm$ 0.001 & 0.054 $\pm$ 0.003 & 0.052 $\pm$ 0.001 & 0.049 $\pm$ 0.001 & 0.051 $\pm$ 0.001 & 0.050 $\pm$ 0.001 & 0.059 $\pm$ 0.001 & 0.053 $\pm$ 0.001 & 0.052 $\pm$ 0.001 & 0.050 $\pm$ 0.001 \\
 & \multirow{2}{*}{Transfer $L_2$} & $\mathcal{R}$ & 0.675 $\pm$ 0.049 & \tblsecond{0.783 $\pm$ 0.043} & \tblthird{0.783 $\pm$ 0.031} & 0.720 $\pm$ 0.039 & 0.764 $\pm$ 0.066 & 0.683 $\pm$ 0.037 & 0.619 $\pm$ 0.044 & 0.526 $\pm$ 0.023 & 0.522 $\pm$ 0.020 & 0.541 $\pm$ 0.018 & 0.686 $\pm$ 0.018 & 0.686 $\pm$ 0.023 & 0.760 $\pm$ 0.019 & \tblbest{0.835 $\pm$ 0.032} \\
 &  & LERS & \tblthird{0.082 $\pm$ 0.006} & \tblbest{0.095 $\pm$ 0.005} & \tblsecond{0.088 $\pm$ 0.003} & 0.066 $\pm$ 0.004 & 0.070 $\pm$ 0.006 & 0.063 $\pm$ 0.003 & 0.057 $\pm$ 0.004 & 0.040 $\pm$ 0.002 & 0.041 $\pm$ 0.002 & 0.043 $\pm$ 0.001 & 0.051 $\pm$ 0.001 & 0.052 $\pm$ 0.002 & 0.059 $\pm$ 0.001 & 0.065 $\pm$ 0.003 \\
 & \multirow{2}{*}{Transfer $L_\infty$} & $\mathcal{R}$ & 0.711 $\pm$ 0.026 & 0.749 $\pm$ 0.024 & 0.782 $\pm$ 0.025 & \tblthird{0.785 $\pm$ 0.014} & 0.778 $\pm$ 0.022 & 0.751 $\pm$ 0.019 & 0.726 $\pm$ 0.022 & 0.650 $\pm$ 0.014 & 0.636 $\pm$ 0.011 & 0.649 $\pm$ 0.011 & 0.741 $\pm$ 0.010 & 0.744 $\pm$ 0.014 & \tblsecond{0.793 $\pm$ 0.015} & \tblbest{0.847 $\pm$ 0.018} \\
 &  & LERS & \tblthird{0.086 $\pm$ 0.003} & \tblbest{0.091 $\pm$ 0.003} & \tblsecond{0.088 $\pm$ 0.003} & 0.072 $\pm$ 0.001 & 0.071 $\pm$ 0.002 & 0.069 $\pm$ 0.002 & 0.067 $\pm$ 0.002 & 0.050 $\pm$ 0.001 & 0.050 $\pm$ 0.001 & 0.052 $\pm$ 0.001 & 0.055 $\pm$ 0.001 & 0.056 $\pm$ 0.001 & 0.061 $\pm$ 0.001 & 0.066 $\pm$ 0.001 \\
 & \multirow{2}{*}{NES} & $\mathcal{R}$ & 0.444 $\pm$ 0.053 & 0.562 $\pm$ 0.088 & 0.614 $\pm$ 0.043 & 0.600 $\pm$ 0.030 & 0.610 $\pm$ 0.059 & 0.634 $\pm$ 0.079 & 0.366 $\pm$ 0.044 & 0.726 $\pm$ 0.026 & 0.681 $\pm$ 0.021 & 0.655 $\pm$ 0.018 & \tblbest{0.803 $\pm$ 0.014} & \tblthird{0.732 $\pm$ 0.033} & \tblsecond{0.784 $\pm$ 0.030} & 0.599 $\pm$ 0.020 \\
 &  & LERS & 0.054 $\pm$ 0.006 & \tblsecond{0.068 $\pm$ 0.011} & \tblbest{0.069 $\pm$ 0.005} & 0.055 $\pm$ 0.003 & 0.056 $\pm$ 0.005 & 0.058 $\pm$ 0.007 & 0.034 $\pm$ 0.004 & 0.056 $\pm$ 0.002 & 0.053 $\pm$ 0.002 & 0.052 $\pm$ 0.001 & 0.060 $\pm$ 0.001 & 0.055 $\pm$ 0.003 & \tblthird{0.061 $\pm$ 0.002} & 0.047 $\pm$ 0.002 \\
 & \multirow{2}{*}{CW} & $\left\lVert \delta_{\mathrm{tex}} \right\rVert_\infty$ & 0.940 $\pm$ 0.039 & 0.966 $\pm$ 0.018 & 0.969 $\pm$ 0.018 & 0.947 $\pm$ 0.027 & 0.968 $\pm$ 0.012 & 0.964 $\pm$ 0.020 & 0.914 $\pm$ 0.043 & \tblsecond{0.980 $\pm$ 0.006} & \tblthird{0.978 $\pm$ 0.009} & {0.975 $\pm$ 0.013} & \tblbest{--$^\dagger$} & 0.836 $\pm$ 0.000 & 0.717 $\pm$ 0.116 & 0.967 $\pm$ 0.010 \\
 &  & $\left\lVert \delta_{\mathrm{tex}} \right\rVert_2$ & 16.10 $\pm$ 2.41 & 22.97 $\pm$ 8.30 & 23.95 $\pm$ 7.95 & 17.14 $\pm$ 5.32 & 20.49 $\pm$ 7.64 & 20.21 $\pm$ 3.78 & 13.77 $\pm$ 3.30 & \tblsecond{26.14 $\pm$ 6.41} & \tblthird{24.87 $\pm$ 9.59} & {24.51 $\pm$ 5.31} & \tblbest{--$^\dagger$} & 4.00 $\pm$ 0.00 & 4.64 $\pm$ 4.20 & 16.35 $\pm$ 0.00 \\
\midrule
\multirow{12}{*}{\rotatebox[origin=c]{90}{\textbf{Common video perturbations}}} & \multirow{2}{*}{AWN (texture)} & $\mathcal{R}$ & 0.989 $\pm$ 0.005 & 0.985 $\pm$ 0.003 & 0.991 $\pm$ 0.004 & \tblsecond{0.993 $\pm$ 0.005} & \tblthird{0.992 $\pm$ 0.007} & 0.931 $\pm$ 0.032 & 0.904 $\pm$ 0.036 & 0.970 $\pm$ 0.008 & 0.976 $\pm$ 0.007 & 0.962 $\pm$ 0.014 & 0.986 $\pm$ 0.005 & 0.991 $\pm$ 0.004 & \tblbest{0.993 $\pm$ 0.003} & 0.914 $\pm$ 0.034 \\
 &  & LERS & \tblbest{0.120 $\pm$ 0.001} & \tblsecond{0.119 $\pm$ 0.000} & \tblthird{0.111 $\pm$ 0.000} & 0.091 $\pm$ 0.000 & 0.091 $\pm$ 0.001 & 0.085 $\pm$ 0.003 & 0.083 $\pm$ 0.003 & 0.075 $\pm$ 0.001 & 0.076 $\pm$ 0.001 & 0.077 $\pm$ 0.001 & 0.073 $\pm$ 0.000 & 0.075 $\pm$ 0.000 & 0.077 $\pm$ 0.000 & 0.071 $\pm$ 0.003 \\
 & \multirow{2}{*}{AWN (frames)} & $\mathcal{R}$ & 0.942 $\pm$ 0.013 & \tblthird{0.957 $\pm$ 0.008} & \tblsecond{0.972 $\pm$ 0.010} & 0.930 $\pm$ 0.013 & 0.896 $\pm$ 0.022 & 0.441 $\pm$ 0.014 & 0.670 $\pm$ 0.023 & 0.103 $\pm$ 0.004 & 0.088 $\pm$ 0.002 & 0.199 $\pm$ 0.018 & \tblbest{0.993 $\pm$ 0.003} & 0.950 $\pm$ 0.012 & 0.921 $\pm$ 0.020 & 0.868 $\pm$ 0.016 \\
 &  & LERS & \tblsecond{0.114 $\pm$ 0.002} & \tblbest{0.116 $\pm$ 0.001} & \tblthird{0.109 $\pm$ 0.001} & 0.086 $\pm$ 0.001 & 0.082 $\pm$ 0.002 & 0.041 $\pm$ 0.001 & 0.062 $\pm$ 0.002 & 0.008 $\pm$ 0.000 & 0.007 $\pm$ 0.000 & 0.016 $\pm$ 0.001 & 0.074 $\pm$ 0.000 & 0.072 $\pm$ 0.001 & 0.071 $\pm$ 0.002 & 0.068 $\pm$ 0.001 \\
 & \multirow{2}{*}{Shot noise} & $\mathcal{R}$ & 0.974 $\pm$ 0.005 & 0.981 $\pm$ 0.003 & \tblsecond{0.985 $\pm$ 0.004} & 0.963 $\pm$ 0.011 & 0.964 $\pm$ 0.009 & 0.685 $\pm$ 0.027 & 0.890 $\pm$ 0.015 & 0.142 $\pm$ 0.009 & 0.153 $\pm$ 0.011 & 0.225 $\pm$ 0.028 & \tblbest{0.995 $\pm$ 0.002} & \tblthird{0.984 $\pm$ 0.002} & 0.979 $\pm$ 0.004 & 0.944 $\pm$ 0.007 \\
 &  & LERS & \tblsecond{0.118 $\pm$ 0.001} & \tblbest{0.119 $\pm$ 0.000} & \tblthird{0.111 $\pm$ 0.000} & 0.089 $\pm$ 0.001 & 0.089 $\pm$ 0.001 & 0.063 $\pm$ 0.003 & 0.082 $\pm$ 0.001 & 0.011 $\pm$ 0.001 & 0.012 $\pm$ 0.001 & 0.018 $\pm$ 0.002 & 0.074 $\pm$ 0.000 & 0.074 $\pm$ 0.000 & 0.076 $\pm$ 0.000 & 0.074 $\pm$ 0.001 \\
 & \multirow{2}{*}{Flicker} & $\mathcal{R}$ & 0.925 $\pm$ 0.015 & 0.880 $\pm$ 0.012 & \tblthird{0.968 $\pm$ 0.010} & 0.871 $\pm$ 0.014 & 0.890 $\pm$ 0.017 & 0.662 $\pm$ 0.018 & 0.708 $\pm$ 0.020 & \tblsecond{0.989 $\pm$ 0.004} & \tblbest{0.992 $\pm$ 0.003} & 0.961 $\pm$ 0.007 & 0.663 $\pm$ 0.028 & 0.420 $\pm$ 0.032 & 0.435 $\pm$ 0.023 & 0.308 $\pm$ 0.013 \\
 &  & LERS & \tblbest{0.112 $\pm$ 0.002} & \tblthird{0.107 $\pm$ 0.001} & \tblsecond{0.109 $\pm$ 0.001} & 0.080 $\pm$ 0.001 & 0.082 $\pm$ 0.002 & 0.061 $\pm$ 0.002 & 0.065 $\pm$ 0.002 & 0.076 $\pm$ 0.000 & 0.077 $\pm$ 0.000 & 0.077 $\pm$ 0.001 & 0.049 $\pm$ 0.002 & 0.032 $\pm$ 0.002 & 0.034 $\pm$ 0.002 & 0.024 $\pm$ 0.001 \\
 & \multirow{2}{*}{Motion jitter} & $\mathcal{R}$ & \tblsecond{0.986 $\pm$ 0.003} & \tblthird{0.984 $\pm$ 0.004} & \tblbest{0.988 $\pm$ 0.004} & 0.646 $\pm$ 0.020 & 0.856 $\pm$ 0.009 & 0.605 $\pm$ 0.016 & 0.626 $\pm$ 0.021 & 0.561 $\pm$ 0.021 & 0.592 $\pm$ 0.018 & 0.585 $\pm$ 0.014 & 0.504 $\pm$ 0.010 & 0.603 $\pm$ 0.011 & 0.555 $\pm$ 0.008 & 0.705 $\pm$ 0.008 \\
 &  & LERS & \tblbest{0.119 $\pm$ 0.000} & \tblsecond{0.119 $\pm$ 0.000} & \tblthird{0.111 $\pm$ 0.000} & 0.060 $\pm$ 0.002 & 0.079 $\pm$ 0.001 & 0.056 $\pm$ 0.002 & 0.057 $\pm$ 0.002 & 0.043 $\pm$ 0.002 & 0.046 $\pm$ 0.001 & 0.047 $\pm$ 0.001 & 0.037 $\pm$ 0.001 & 0.046 $\pm$ 0.001 & 0.043 $\pm$ 0.001 & 0.055 $\pm$ 0.001 \\
 & \multirow{2}{*}{Fog} & $\mathcal{R}$ & 0.875 $\pm$ 0.018 & 0.955 $\pm$ 0.010 & 0.928 $\pm$ 0.010 & 0.943 $\pm$ 0.013 & 0.929 $\pm$ 0.013 & 0.810 $\pm$ 0.024 & 0.782 $\pm$ 0.028 & 0.975 $\pm$ 0.002 & 0.977 $\pm$ 0.001 & 0.954 $\pm$ 0.009 & \tblthird{0.984 $\pm$ 0.005} & \tblsecond{0.997 $\pm$ 0.001} & \tblbest{0.999 $\pm$ 0.000} & 0.925 $\pm$ 0.014 \\
 &  & LERS & \tblsecond{0.106 $\pm$ 0.002} & \tblbest{0.116 $\pm$ 0.001} & \tblthird{0.104 $\pm$ 0.001} & 0.087 $\pm$ 0.001 & 0.085 $\pm$ 0.001 & 0.074 $\pm$ 0.002 & 0.072 $\pm$ 0.003 & 0.075 $\pm$ 0.000 & 0.076 $\pm$ 0.000 & 0.076 $\pm$ 0.001 & 0.073 $\pm$ 0.000 & 0.075 $\pm$ 0.000 & 0.077 $\pm$ 0.000 & 0.072 $\pm$ 0.001 \\
\bottomrule
\end{tabular}%
}
\caption{Robustness $\mathcal{R}$ and log-efficiency robustness score (LERS) across networks and perturbation types. Models are grouped by architecture family; the vCNN variants use 2-, 5-, and 10-frame temporal contexts, respectively. hexRAFT denotes the FlyNet-matched RAFT variants. For each row, the best-performing mean is bold and underlined, the second-best is bold, and the third-best is italicized; the corresponding uncertainty receives the same formatting. Tied values share the same rank. The smallest GFLOPs value is highlighted in bold. The RAFT family is from \citep{RAFT2020}, and FlyNet from \citep{Lappalainen2024}. For CW, we report the mean and standard deviation of the minimum perturbation magnitude across runs attaining at least 80\% misclassified frames. Empty cells indicate missing results; $^\dagger$ indicates that CW data are present but no qualifying run was found.}
\label{table:robustness_full}
\end{table*}
\end{landscape}


\end{document}